\documentclass[lettersize,journal,onecolumn]{IEEEtran}
\usepackage{amsmath,amsfonts}
\usepackage{algorithmic}
\usepackage{algorithm}
\usepackage{algorithmic}
\usepackage{array}
\usepackage{textcomp}
\usepackage{stfloats}
\usepackage{url}
\usepackage{verbatim}
\usepackage{graphicx}
\usepackage{cite}
\usepackage{siunitx} 
\usepackage{caption}
\usepackage{xcolor}
\usepackage[caption=false,font=normalsize,labelfont=sf,textfont=sf]{subfig}
\usepackage{stfloats}
\usepackage{multirow}
\usepackage[version=4]{mhchem} 
\begin{document}

\title{An Efficient and Accurate Memristive Memory for Array-based Spiking Neural Networks}

\author{Hritom Das,\IEEEmembership{ Member,~IEEE}, Rocco D. Febbo, S.N.B. Tushar,\IEEEmembership{ Graduate Student Member,~IEEE}, Nishith N. Chakraborty,\IEEEmembership{ Graduate Student Member,~IEEE}, Maximilian Liehr,\IEEEmembership{ Member,~IEEE}, \\Nathaniel C. Cady,\IEEEmembership{ Member,~IEEE}, and Garrett S. Rose,~\IEEEmembership{Senior Member,~IEEE}
\thanks{Manuscript received 20 March 2023; revised 30 June 2023; accepted 24 July 2023. This work was supported by Air Force Research Laboratory under agreement numbers FA8750-21-1-1018 and FA8750-21-1-1019.  The U.S. Government may reproduce and distribute reprints for Governmental purposes, despite any copyright notation. The views and conclusions expressed herein are solely those of the authors and do not necessarily reflect the official policies or endorsements of the Air Force Research Laboratory or the U.S. Government. 

(Hritom Das and Rocco D. Febbo contributed equally to this work.) (Corresponding author: Hritom Das.)\\
Hritom Das, Rocco D. Febbo, S.N.B. Tushar, Nishith N. Chakraborty, and Garrett S. Rose  are with the Department of Electrical Engineering and Computer Science, University of Tennessee, Knoxville, TN 37996, USA (e-mail: hdas, rfebbo, stushar1, nchakra1, garose@utk.edu) \\
Maximilian Liehr and  Nathaniel Cady are with the Department of Nanoscale Science \& Engineering, SUNY Polytechnic Institute, Albany, NY 12203, USA (e-mail: liehrmw, cadyn@sunypoly.edu)
}
\thanks{}
}

\markboth{}%
{Shell \MakeLowercase{\textit{et al.}}: A Sample Article Using IEEEtran.cls for IEEE Journals}


\maketitle

\begin{abstract}
Memristors provide a tempting solution for weighted synapse connections in neuromorphic computing due to their size and non-volatile nature. However, memristors are unreliable in the commonly used voltage-pulse-based programming approaches and require precisely shaped pulses to avoid programming failure. In this paper, we demonstrate a current-limiting-based solution that provides a more predictable analog memory behavior when reading and writing memristive synapses. With our proposed design READ current can be optimized by $\sim$19x compared to the 1T1R design. Moreover, our proposed design saves $\sim$9x energy compared to the 1T1R design. Our 3T1R design also shows promising write operation which is less affected by the process variation in MOSFETs and the inherent stochastic behavior of memristors. Memristors used for testing are hafnium oxide based and were fabricated in a \SI{65}{\nano\meter} hybrid CMOS-memristor process. The proposed design also shows linear characteristics between the voltage applied and the resulting resistance for the writing operation. The simulation and measured data show similar patterns with respect to voltage pulse based programming and current compliance based programming. We further observed the impact of this behavior on neuromorphic-specific applications such as a spiking neural network.
\end{abstract}


\section{Introduction}
\IEEEPARstart{O}{ver} the last decade, artificial intelligence has added a useful suite of new computational tools. However, utilizing these tools in an effective manner with limited access to internet and power has proven to be a challenge. Meeting such requirements, using state-of-the-art computing systems, is hindered in part by the separation of the memory and computation unit \cite{b1}, which causes large energy consumption during the constant rapid transfer of data between memory and processor. This problem can be mitigated by designing a circuit which uses in-memory computing \cite{b2, IM1, IM2, IM3}. A classical example of in-memory computing is the dot product engine (DPE) \cite{b3, b4, DPE2}, which can be the backbone of many machine learning algorithms including neuromorphic computing \cite{b10, b11} and reservoir computing \cite{RC1, RC2}. The most basic implementation of a DPE performs a dot product between a vector and a matrix. The energy during this operation can be reduced further if a memory device such as a memristor is used to store the values of the matrix as a function of their stored resistance. Due to lower \textit{READ} power and multi-level storage capacity, the memristor device shows a higher energy efficiency compared to conventional digital memory such as SRAM or DRAM\cite{b1, dm1, dm2, dm3}.  

A memristor is a two-terminal device that can act as an analog memory\cite{R1}. By applying a specific voltage to the top electrode for some time, the memristor can have its resistance value lowered through the formation of a filament through the material. This is known as the low resistance state (LRS). By applying a negative voltage to the top electrode of the memristor, this filament can be broken which RESETs the memristor into a high resistance state (HRS). For example, the memristive devices used in this paper (resistive random access memory, or RRAM in a one transistor one RRAM or 1T1R configuration) utilize hafnium oxide as the switching material and can be \textit{SET} into a LRS by applying a voltage of positive \SI{0.7}{\volt}. The purpose of the \textit{SET} operation is analogous to writing data to memory. During this process, a specific final resistance value is targeted which will later be sensed during \textit{READ} operations. During this \textit{SET} process, the memristor's resistance will rapidly reduce as long as the voltage is above the threshold. The lowering of the resistance reaches a stopping point once the voltage goes below the threshold. On the other hand, if a constant current is maintained, this stopping point will occur at a point relative to the magnitude of the current applied. This is due to Ohm's law. Therefore, by limiting the maximum current, a minimum memristor resistance value can be targeted.

When designing memristive based memory systems, specific challenges need to be considered.
Managing variation across devices and during read\slash write cycles is one of these challenges for analog memories. The behavior is required to be predictable across all structures in order to have a reliable memory. This variation can cause large discrepancies between operation during simulation and post-fabrication. In memristive memory, this is especially apparent due to the inherent stochastic behavior in the memristors themselves. This is largely due to the unpredictable process of filament formation which takes place during a SET operation. To minimize these effects, an LRS range of  5 k$\Omega$ to 20 k$\Omega$ is utilized due to its more predictable behavior compared to HRS.
Another consideration is the voltage pulse shape during programming \cite{reliability1, reliability2, reliability3}. If the pulse shaping is not precise, the write resistance values can vary from their targeted value. To overcome these issues, the 3T1R memristive circuit can be utilized. This circuit is programmed utilizing current compliance through the memristor as opposed to a voltage pulse across the memristor. Also, by targeting the current through the memristor, a low-power array-based spiking neural network can be designed. If we consider a DPE design, the READ power is more crucial than the write power. This is due to the memory element only being written during the initial training operation. After training, the system will be READ repeatedly during the testing process and after deployment. Therefore, the overall power consumption of a design such as a DPE can be optimized if the READ current is reduced.

Analog implementations of a DPE using memristors as the analog storage elements shows promising features for several application domains \cite{b4, DPE1, DPE2}. However, due to the inherent variability in memristor operation as discussed earlier, computation will be inaccurate compared to traditional digital implementations \cite{b1}. This variability is due to the stochastic nature in which the memristor forms and breaks its filament. Careful design of the DPE can help mitigate some of this variability. DPE circuits are commonly implemented in a 2D crossbar array configuration where the top of each memristor is connected across the rows. Then the bottom of each memristor is connected to an n-channel MOSFET which is also connected across the columns as seen in Figure \ref{fig:In-line network}. This is known as an in-line configuration.

A memristive DPE can be used in neuromorphic applications such as a spiking neural network (SNN). SNNs perform computations using spikes. These spikes can be applied to a DPE in order to perform a multiplication \cite{RC1, RC2, DPE1, DPE3}. The result of this multiplication can be tailored to a specific application such as a classification problem or a control problem \cite{app1}. In order to get the best performance in the application, the result of the multiplication needs to be predictable within a certain range. 

In this work, toward resolving the above challenges, we make the following contributions:
\begin{enumerate}
  \item   Proposed 3T1R memristive array utilizing current compliance for better reliability. 3T1R configuration provides better controllability during the \textit{SET} and \textit{READ} operations.
  \item   Lower \textit{READ} current for each resistance level is demonstrated. Due to optimized \textit{READ} operation, 3T1R shows lower power consumption compared to 1T1R configuration. 
  \item   Experimental measurements and simulated results show a similar trend for variation in LRS.
  \item     The effect of pulse shaping during programming is observed in simulation and measured data.
  \item   Analog DPE is modeled and simulated to evaluate the reliability and performance of the proposed design. 
  \item  The effect of our proposed accurate and efficient approach for neuromorphic applications are analyzed.
\end{enumerate}


The remainder of the paper is organized as follows. Section II focuses on in-line style memristive DPE design. Here, the READ and write operation of the design are discussed. Section III shows the proposed design with 3T1R structure. Section IV analyses the non-linear regions of the simulation results for SET current and READ current and their impact reliability. In addition, the impact of process variation on both designs is examined using Monte Carlo. Simulated results from cadence using the memristor model from \cite{model_glsvlsi} and our mathematical model is evaluated. In Section V, physical data from a fabricated chip is also analyzed to observe the reliability. Applications are evaluated in Section VI. There are four different classification tasks taken under consideration for an SNN application. A brief overview of peripheral circuity is provided in section VII. Next, a comparison with prior work is discussed in section VIII. Finally, the paper concludes with future work in Section IX.

\section{In-line Style Memristive DPE}
In our proposed design a TiN TE-HfO2-TiN BE material stack is utilized to construct the memristor device. After fabrication with the 65nm CMOS 10LPe process \cite{65}, 26 devices are measured with our probe station to consider process variation and cycle-cycle variation.  A Verilog-A model is derived based on our measured data such as the I-V curve, threshold voltage for switching, switching time, and process variation at the different memristive regions. In addition, curve fitting parameters are also used to better match model performance with measured results. A brief illustration of our model is presented in prior work \cite{model_glsvlsi}. This model is utilized to characterize the memristor devices during a system-level simulation in Cadence Spectre.

\begin{figure}[t]
  \centering
  \includegraphics[width=3.4in]{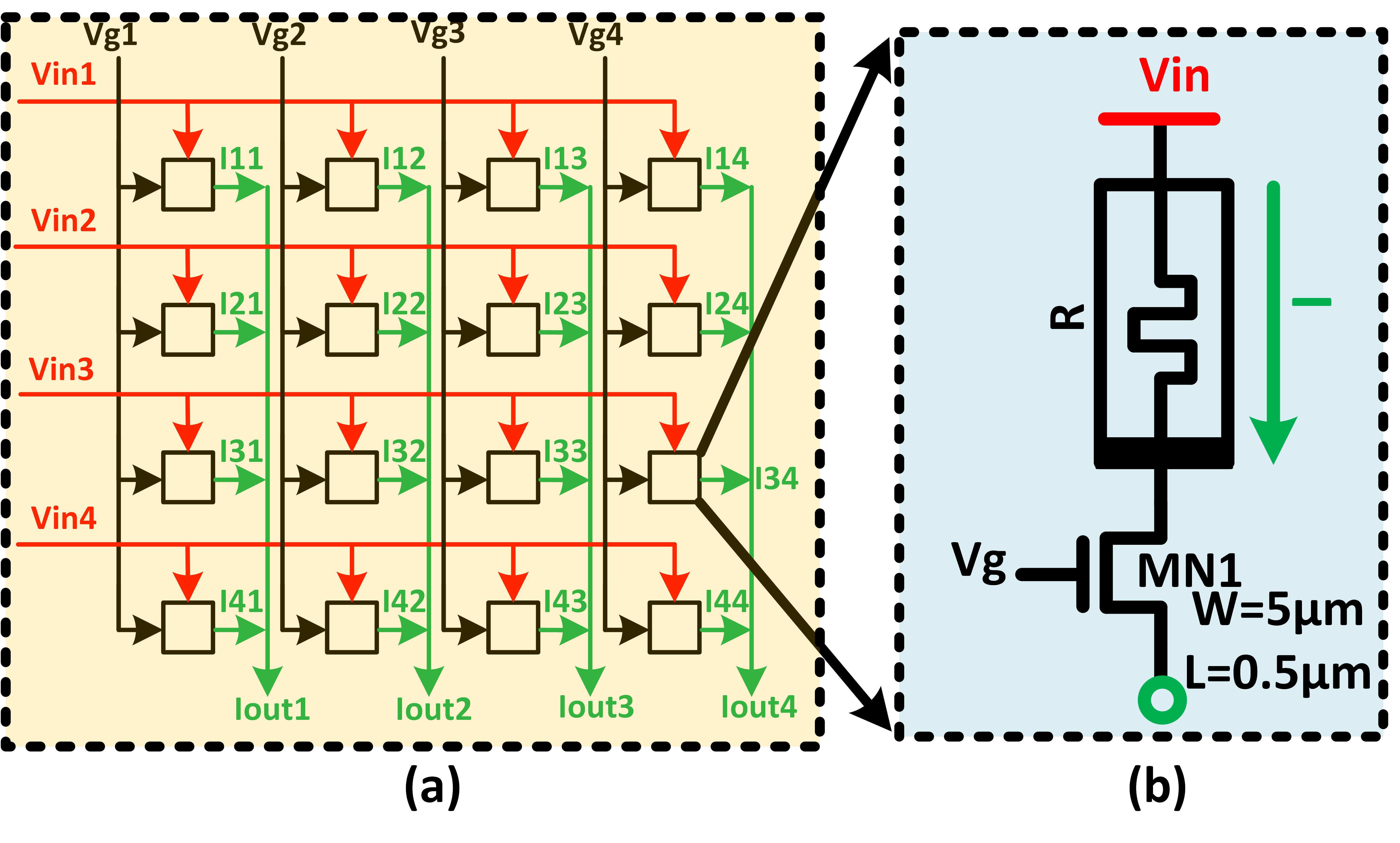}
  \caption{ (a) 4x4 in-line configuration with 1T1R structure. (b) $V_{in}$ (red) is the common connection for all memristors in a row. $V_g$ (black) are the controlling signals (READ or write) for all the memristors in a column. $I_{out}$ (green) is the cumulative current output from each column. The width and length of $MN1$ are \SI{5}{\micro\meter} and \SI{0.5}{\micro\meter} respectively. Here, large thick oxide transistors are required due to the high voltage during forming.}
  \label{fig:In-line network}
\end{figure}

DPEs are very popular for machine learning hardware. There are different architectures available for DPE such as 1T1R (1T: one transistor, 1R: one memristor) \cite{b4}, \cite{DPE2}, \cite{b12}, 3T1R\cite{b7}, and so on. Due to its simple structure and operation, most of the existing designs use the 1T1R configuration \cite{b4},\cite{DPE2}. In this 2D crossbar array configuration, the top of each memristor is connected across the rows. Then the bottom of each memristor is connected to an n-channel MOSFET which is connected across the columns as seen in Figure \ref{fig:In-line network} (a). This describes an in-line configuration.

\subsection{SET \slash Write Operation}

 Fig. \ref{fig:In-line network} (a) illustrates a 4x4 DPE configuration where each cell represents a 1T1R synaptic memory element. There are two input signals, $V_{in}$ and $V_{g}$, for each row and column respectively. Each $V_{in}$ provides a row-wise input pulse across the DPE unit. Moreover, $V_{g}$ provides a column-wise READ or write access across the DPE unit.
 Fig. \ref{fig:In-line network} (b) shows the internal circuit connections of a single 1T1R cell in this DPE. The top node of the memristor is connected to $V_{in}$ with the bottom connected to the drain of the ${NMOS}$ device.  
 Additionally, $V_{g}$ is used as a selector for both READ and write operations in a column. During a write, a voltage pulse is driven across the selected memristor using the $V_{in}$ (top of memristor). This voltage pulse must be carefully shaped in order to drive the appropriate analog voltage for a determined amount of time to properly write the resistance of the device\cite{b10}. Shaping this voltage pulse to produce a reliable write is a significant challenge for the 1T1R configuration. 
 The amplitude of $V_{in}$ for a write operation is in the range of \SI{0.9}{\volt} to \SI{1.1}{\volt}, depending on the value to be written. $V_{g}$ signals work as a digitally controlled signal to access the various cells. 
 Analog values are considered in the LRS range of 5 k$\Omega$ to 20 k$\Omega$ due to more predictable resistance as compared to HRS\cite{b8}.

\subsection{READ Operation}

 During a READ operation, currents flowing through multiple 1T1R cells accumulate into each column of the DPE array to form the output denoted as $I_{out}$. The applied $V_{in}$ could be an analog voltage corresponding to an element in a vector input. Similarly, the resistance of the memristor is expected to have been written with a value corresponding to a matrix element. Thus, the current through a particular 1T1R cell corresponds to the product of a vector element ($V_{in}$) and the inverse of the memristor's resistance. 
 The source of each 1T1R $NMOS$ is shorted with the columnar $I_{out}$ such that its current into the column to be summed with other such currents courtesy Kirchoff's current law. In this case, the $I_{out}$ values of the columns correspond to the output vector elements. $V_{g}$ is the gate signal of the $NMOS$ and it also acts as an access transistor for the READ operation. Due to high voltage forming operation, thick oxide transistors are often required for $NMOS$ and $PMOS$. Moreover, a thick oxide transistor is useful to reduce the flicker noise from the system \cite{b13}. For the READ operation, the $V_{in}$ and $V_{g}$ will be \SI{0.6}{\volt} and \SI{1.2}{\volt}, respectively, to achieve low-power READ operation. All the $V_g$s can be considered as common access signals for DPE operation. As a result, all the column current can be read by activating a single $V_g$. On the other hand, to get more controllability, $V_g$'s can be operated separately and READ as many column currents as needed.  

\section{Proposed 3T1R Style DPE}

\begin{figure}[t]
  \centering
  \includegraphics[width=3.4in]{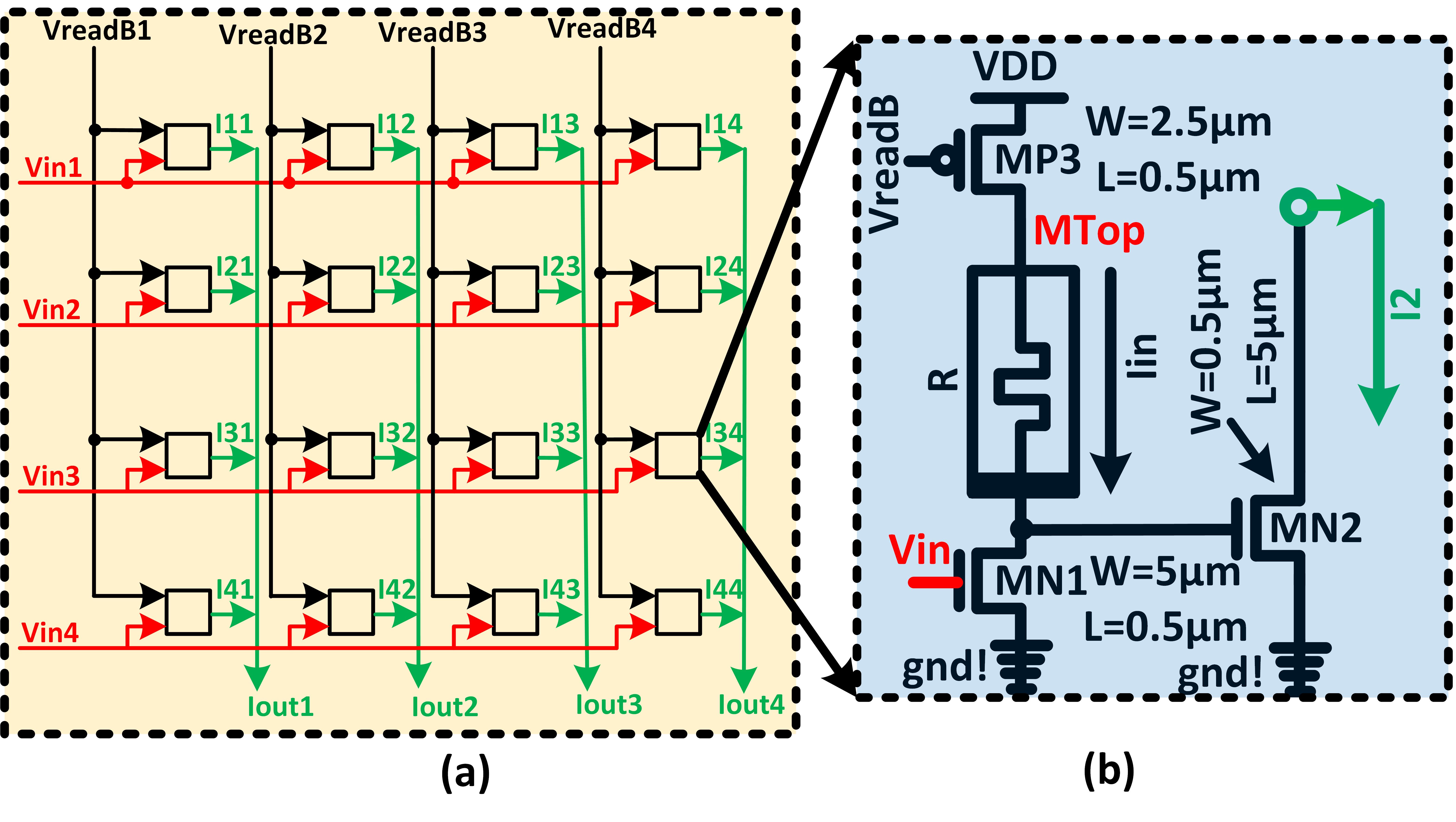}
  \caption{ (a) 4x4 3T1R style DPE configuration is illustrated. (b) $V_{readB}$ is the common connection for all the gate of $PMOS$ in a column. $MP3$ will be "OFF" during a hold operation otherwise, it will be "ON" during a write or READ operation. $V_{in}$ is the controlling signal to access the memristor in a row to do READ and write operations. The $V_{in}$ is from \SI{0.8}{\volt} to \SI{1.1}{\volt}, during a write operation. On the other hand, the $V_{in}$ is \SI{0.6}{\volt}, during a READ operation. $I_{out}$ is the cumulative current output from a specific column. Here, the drain of each $MN2$ will be shorted together for a DPE configuration. Each cell contains its own ground connection for the source of MN1 and MN2. 
}
  \label{fig:3T1R}
\end{figure}

Fig. \ref{fig:3T1R} shows the proposed 3T1R design for memristive DPEs. Our proposed DPE utilizes a current-controlled memristive synapse as the base memory cell. Fig. \ref{fig:3T1R} (a) illustrates a 4x4 DPE arrangement with our current-controlled synapse. According to Fig. \ref{fig:3T1R} (b) $V_{readB}$ and $V_{in}$ are the input signals for each 3T1R cell. Similar to the in-line configuration, $I_{out}$ are the column currents representing output vector elements in the DPE circuitry. Fig. \ref{fig:3T1R} (b) shows the internal configuration of the 3T1R cell. In this circuit, two $NMOS$ transistors, one $PMOS$ transistor, and a memristor (R) are used. In the first stage, the drain of $MP3$ is connected to the top of the memristor and the drain of $MN1$ is connected at the bottom of the memristor. The presence of the $PMOS$ helps reduce leakage current through the column. For low-power design, the memristor can also be isolated from $VDD$ when the circuit is not in operation (hold operation). On the other hand, in 1T1R combination, the top node of the memristor is always shorted with $V_{in}$. Thus, additional control circuitry is required. $MN2$ plays an important role to obtain a reduced \textit{READ} current for our proposed synapse. The gate of the $MN2$ is controlled precisely with the help of a gate voltage that is generated by the product of the resistance of the memristor and the $1^{st}$ stage current (Iin).

\subsection{SET \slash Write Operation}

In Fig. \ref{fig:3T1R} (b), the input current $I_{in}$ is controlled using the $V_{in}$ at the gate of $MN1$. 
This current will only depend on $V_{in}$ with $MN1$ acting as a current sink for the circuit. 
For a SET or write operation, $V_{readB}$ is used to digitally turned ``ON'' $MP3$. The input $V_{in}$ is driven in the range from \SI{0.8}{\volt} to \SI{1.1}{\volt} to SET the resistance level in the range from 5 k$\Omega$ to 20 k$\Omega$. 

\subsection{READ Operation}

After a SET operation, the analog value of the 3T1R cell is also READ using a current $I_{in}$ controlled by $MN1$.
For this, $V_{readB}$ will set $MP3$ ``ON'' with \SI{0}{\volt} and $V_{in}$ will set $MN1$ into the linear region of operation with \SI{0.6}{\volt}. To control the current and reduce power consumption, $MN1$ is operated at a lower voltage. The voltage at the bottom of the memristor will turn on the $MN2$ with a minimal gate voltage. Finally, the READ current $I_2$ is measured from the drain of the $MN2$. This cell-level output current $I_2$ is connected to the common column to be summed with currents from other cells, again leveraging KCL. In the next section, an empirical comparison between 1T1R and 3T1R is provided.

\section{Cadence Simulation Results}
\subsection{SET and READ Simulation Results}
\begin{figure}[t]
            \centering
            \includegraphics[width=3.4in]{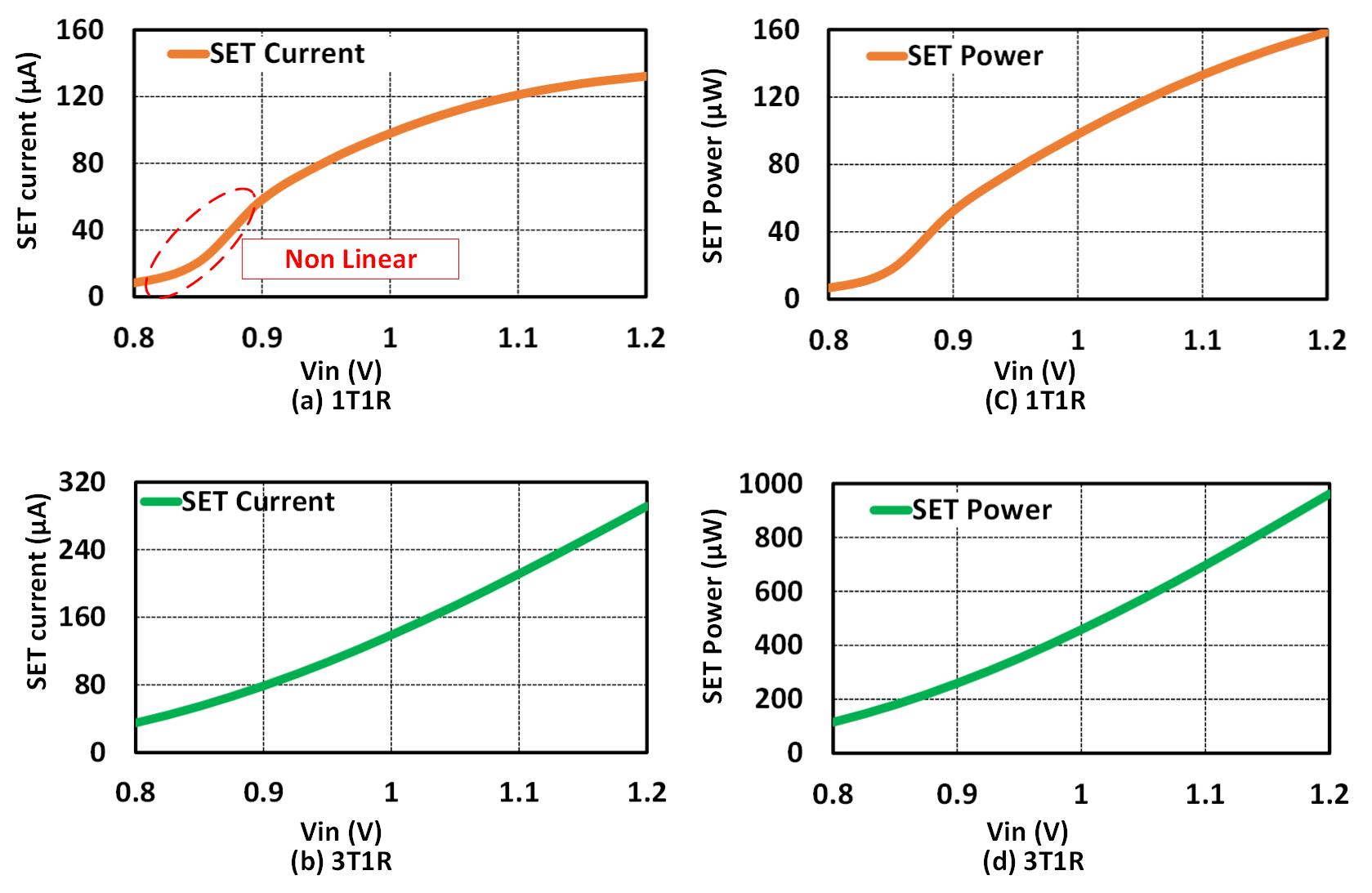}
            \caption{ Cadence simulations are run with \SI{65}{\nano\meter} CMOS technology at room temperature. (a) shows the simulated result for 1T1R device during SET operation. SET voltages are varied from \SI{0.8}{\volt} to \SI{1.2}{\volt}. (b) illustrates the SET current simulation for 3T1R configuration. (c) shows the power simulation for 1T1R device. (d) plots the power range for the 3T1R combination.}
            \label{fig:IP}
        \end{figure}

In our simulation, a \SI{65}{\nano\meter} CMOS 10LPe process from IBM is utilized \cite{65}. A Verilog-A model is used to characterize the memristive device \cite{model_glsvlsi}. A statistical block in this model shows the process variation and stochastic behavior of the memristor. Monte Carlo analysis is used to observe the effect of process variation and non-linear effects of the 1T1R and 3T1R circuits. Fig. \ref{fig:IP} (a) and (b) show the simulation results for SET operation with an initial resistance of 100 k$\Omega$ at different $V_{in}$. The range is varied from \SI{0.8}{\volt} to \SI{1.2}{\volt}. The SET current for 1T1R cell varies from a low current of about \SI{7}{\micro\ampere} to \SI{132}{\micro\ampere}. On the other hand, the SET current of 3T1R is slightly higher and varies from about \SI{35}{\micro\ampere} to \SI{292}{\micro\ampere}. Here, the high SET current of 3T1R is a trade-off for better accessibility for write operations. 3T1R has a more desirable writing functionality compared to 1T1R since the SET current is more linear with respect to the voltage. Due to Ohms law, the final resistance achieved is directly proportional to the current applied during a SET operation. So, having a linear set current will offer better predictability and reliability for programming operations.  Fig. \ref{fig:IP} (c) and (d), show the power consumption of 1T1R and 3T1R cells respectively. Due to the high SET current, 3T1R draws higher power than 1T1R devices. According to Fig. \ref{fig:IP} (c), a SET operation of 1T1R shows the power range from  about \SI{5}{\micro\watt} to \SI{160}{\micro\watt}. On the other hand, the power consumption of 3T1R is significantly higher, ranging from about \SI{115}{\micro\watt} to \SI{964}{\micro\watt}. SET current or power is only considered when the memory array will be utilized for training the dataset. READ power has a higher impact than write power on the total power consumption of a system such as a DPE. However, writing a precise resistance value to the memristor is critical for repeatable results in machine learning.

\begin{figure}[t]
  \centering
  \includegraphics[width=3.4in]{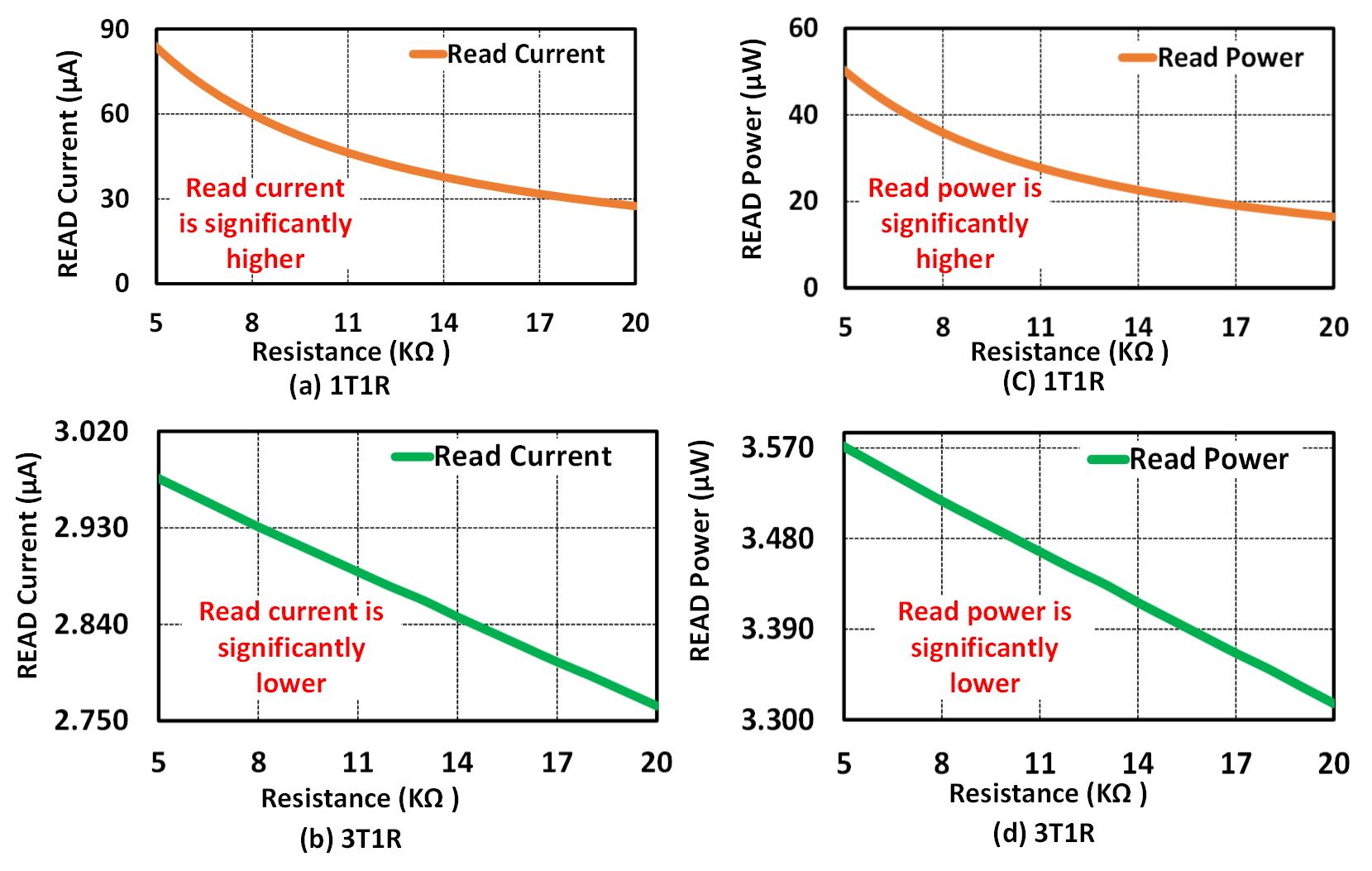}
  \caption{ Simulation of the current during the READ operation for both devices are shown here. (a) shows the READ current for 1T1R devices from 5 k$\Omega$ to 20 k$\Omega$. (b) illustrated the READ current simulation for 3T1R devices for the same resistance range. 3T1R shows a more linear and lower current level according to the resistance level. (c) and (d) show the power consumption of the 1T1R and 3T1R circuits, respectively.}
  \label{fig:READ_IP}
\end{figure}

In the READ operation, it is important to precisely represent the weight value for any machine learning application. Thus, the usability of the DPE design hinges on if the current values for a specific resistance level are distinguishable or not. At first, we consider the 1T1R combination and READ the output current $I_2$ when the memristor resistance is varied in the range from 5 k$\Omega$ to 20 k$\Omega$. According to Fig. \ref{fig:READ_IP} (a), the READ current range is non-linear for the 1T1R, and ranges from about \SI{110}{\micro\ampere} to \SI{35}{\micro\ampere}. Due to the non-linearity, the current values are not easily distinguishable at the specific resistance levels. As a result, weight precision is limited in this design. On the other hand, our proposed design can differentiate resistance levels from 5 k$\Omega$ to 20 k$\Omega$ due to improved linearity. As seen in Fig. \ref{fig:READ_IP} (b), the 3T1R design was also able to achieve a lower current and thus lower power. This linear output stage makes our design suitable for higher resolution weights, as compared to the 1T1R. Here, the READ current range for 3T1R is from about \SI{3.005}{\micro\ampere} to \SI{2.833}{\micro\ampere}. Fig. \ref{fig:READ_IP} (c) and (d) illustrate the corresponding power consumption for 1T1R and 3T1R respectively. At 5 k$\Omega$ and 20 k$\Omega$, we can save up to about 14x power with our proposed design as compared to 1T1R during the READ operation. It is worth noting that currents from only the final stage is considered for power calculations. However, in Section VII power from both stages is considered for the 3T1R design.

\begin{figure}[t]
  \centering
  \includegraphics[width=3.4in]{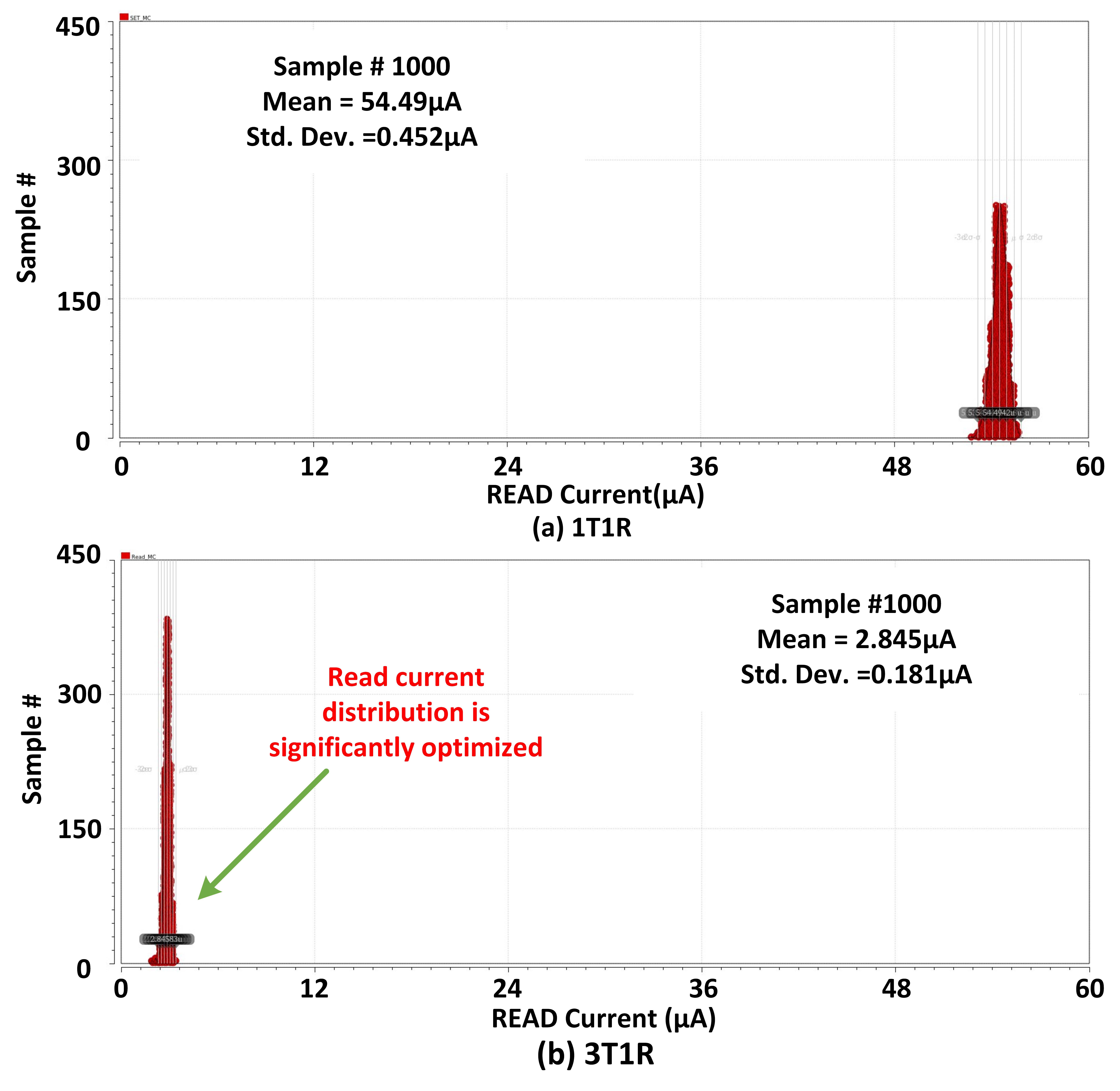}
  \caption{ Monte Carlo simulation has been taken under consideration to analyze the effect of process variation on READ current. 1000 samples are utilized for process variation analysis with Cadence Spectre. (a) shows the Monte Carlo simulation of the 1T1R device during a READ operation at \SI{0.6}{\volt} and 9 k$\Omega$ (b) exhibits the Monte Carlo analysis for the 3T1R device for the same resistance and voltage levels.}
  \label{fig:READ_MC}
\end{figure}

\subsection{Monte Carlo Analysis}

Monte Carlo simulations are a well-known approach to analyze the effects of process variation on a system. Two test cases are analyzed here. At first, the 1T1R device is used for the Monte Carlo simulation. \SI{0.6}{\volt} is applied at $V_{in}$ and the memristor resistance is 9 k$\Omega$. According to Fig. \ref{fig:READ_MC} (a), the mean and std. dev. of the analysis are about \SI{54.5}{\micro\ampere} and \SI{0.452}{\micro\ampere}, respectively. The 3T1R circuit shows a tighter Monte Carlo distribution of READ current. Fig. \ref{fig:READ_MC} (b) shows the Monte Carlo analysis plot of the READ-out current $I_2$ for the 3T1R device. The mean and std. dev. are \SI{2.85}{\micro\ampere} and \SI{0.181}{\micro\ampere} accordingly. Here, the std. dev. of the READ current is lower than the 1T1R device. Moreover, the mean current is significantly lower to READ the same resistance level using our proposed design. Moreover, a range of memristive process variations from 9\% to 31\% is considered for the proposed design. At HRS the process variation is very high and at LRS the variation is lower. A detailed analysis is presented in our prior work\cite{jetcas}.

\subsection{Pulse Shape Effect Based on Cadence Simulation}

\begin{figure}[t]
  \centering
  \includegraphics[width=3.4in]{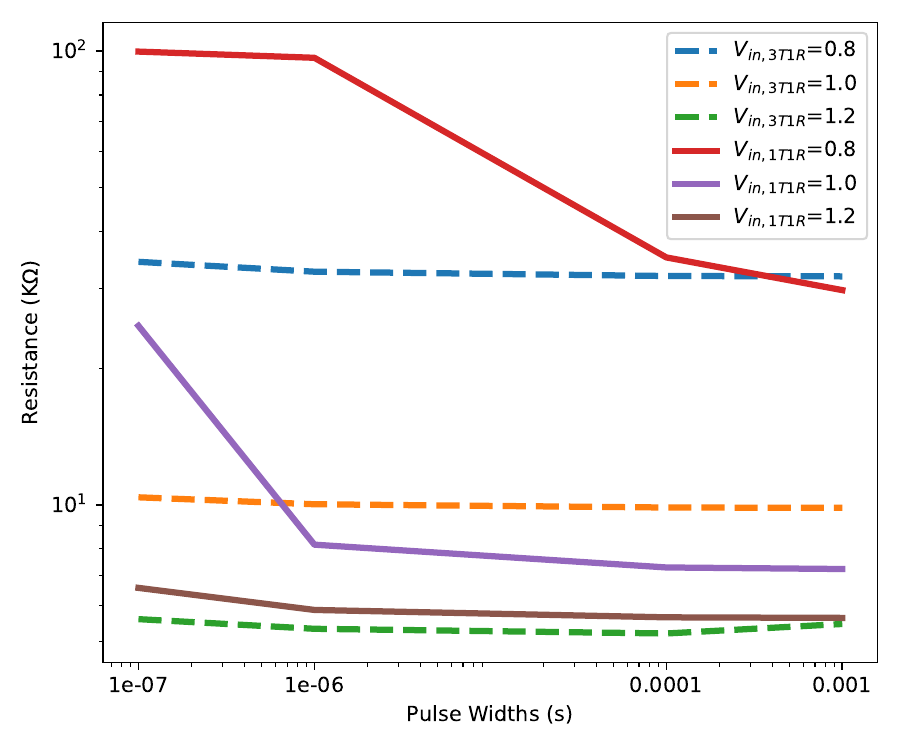}
  \caption{Simulation results of the sensitivity to the pulse shape on 1T1R (solid line) and 3T1R (dotted line) device. This plot illustrates SET operation at different $V_{in}$ such as \SI{0.8}{\volt}, \SI{1}{\volt}, and \SI{1.2}{\volt} . In addition, the pulse width is varied from \SI{100}{\nano\s} to \SI{1}{\milli\s}.  1T1R shows a higher sensitivity to pulse shaping at a lower SET voltage (\SI{0.8}{\volt}) compared to a higher SET voltage (\SI{1.2}{\volt}). On the other hand, 3T1R shows a negligible pulse width effect on the SET operation at any SET voltage. Lower SET voltage is important to keep the design in a low-power region. 3T1R gives us the opportunity to make our design in low-power regions with more reliable SET operation.}
  \label{fig:pulse_cadence_1T1R}
\end{figure}

Fig. \ref{fig:pulse_cadence_1T1R} shows the simulation results on Cadence Spectre at different SET voltages. 1T1R shows a higher pulse width effect on SET operation at a lower SET voltage ($V_{in}$). 
When the SET voltage is \SI{0.8}{\volt}, the SET resistance varies from $\sim$ \SI{100}{\kilo\ohm} to $\sim$ \SI{30}{\kilo\ohm} with pulse width variation from \SI{100}{\nano\s} to \SI{1}{\milli\s}. Due to that, the pulse shape needs to be very specific for the 1T1R device to SET in a specific resistance level. However, a higher SET voltage like \SI{1.2}{\volt} shows lower SET resistance variation at different pulse widths. Alternatively, 3T1R shows negligible variation at any SET voltage with different pulse widths. The 3T1R circuit has less sensitivity due to the addition of $M_{P3}$ and current control from the gate of $M_{N1}$. These results will be verified in the next section using measured data. 

\subsection{Area Overhead Between 1T1R and 3T1R}

\begin{figure}[t]
  \centering
  \includegraphics[width=3.4in]{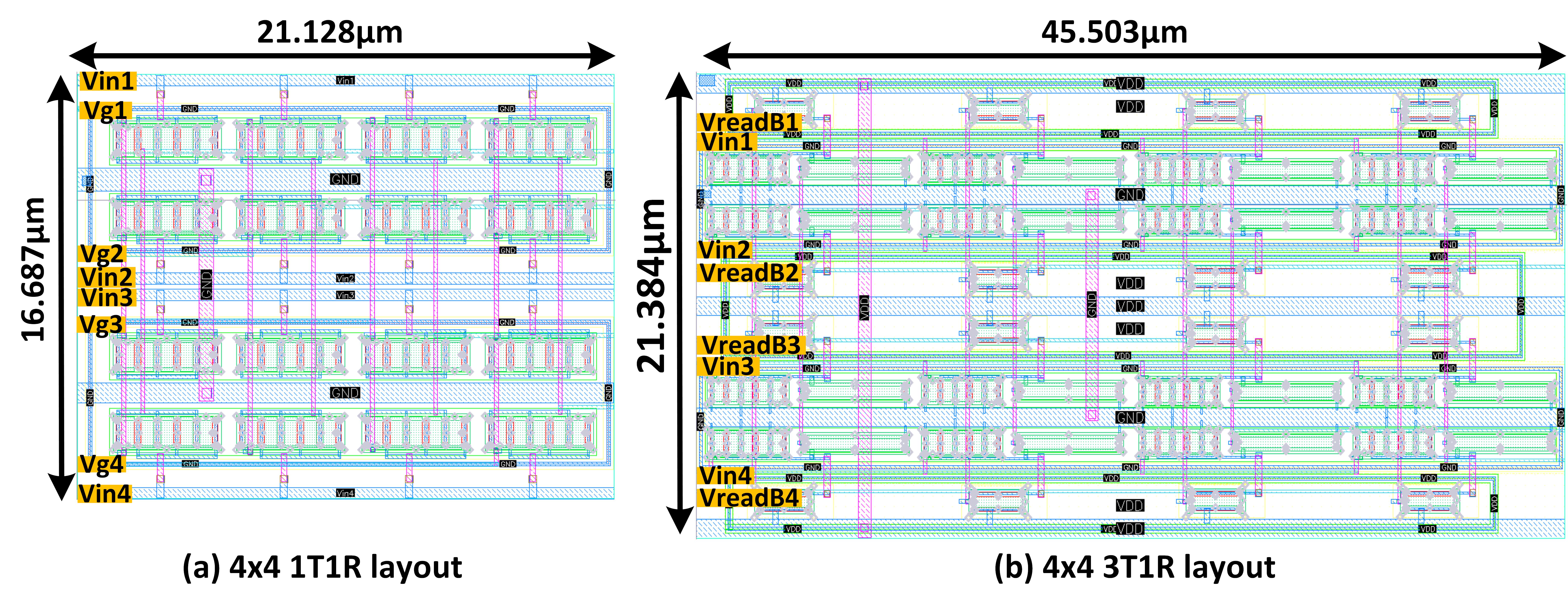}
  \caption{Layout of 1T1R and 3T1R memristive memory array on \SI{65}{\nano\meter} CMOS process. (a) shows a 4x4 memory array of 1T1R cells. (b) illustrates a 4x4 memory array of 3T1R cells. Here, the memristive layer is \ce{HfO2}. }
  \label{fig:layout}
\end{figure}

Fig. \ref{fig:layout} exhibits the layout of the two different design approaches. Fig. \ref{fig:layout} (a) shows the layout of a 4x4 memory array of 1T1R memristive cells, where the area is $\sim$ \SI{353}{\micro\meter^2}. Fig. \ref{fig:layout} (b) illustrates the area of the 4x4 array for 3T1R is $\sim$ \SI{973}{\micro\meter^2}. Due to an extra $NMOS$ and $PMOS$ the area of the 3T1R array is $\sim$2.75x larger than the 1T1R array. This area overhead is a trade-off for improved reliability and power consumption.  

\begin{figure}[t]
  \centering
  \includegraphics[width=2.5in]{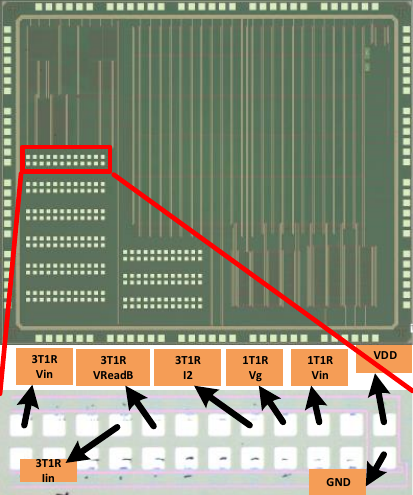}
  \caption{Die photo of our fabricated chip (top). There are five different test structures. We have explored one test structure here. The bottom part is the one 12x2 test structure, where we have 1T1R and 3T1R devices. All the pads are labeled accordingly.}
  \label{fig:chip}
\end{figure}

\section{Measured Data}

\begin{figure}[]
  \centering
  \includegraphics[width=3.4in]{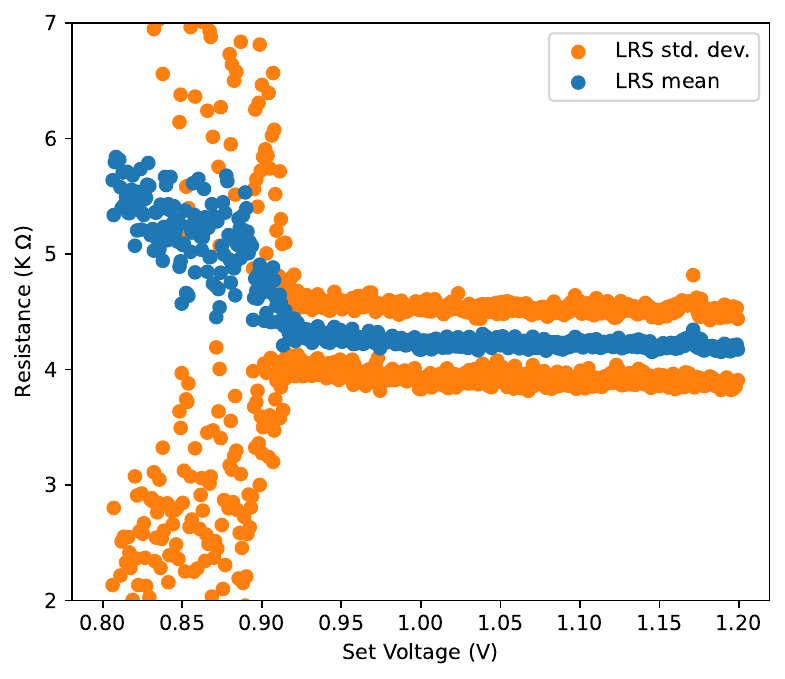}
  \caption{Measured DC test data of 1T1R configuration where gate voltage of NMOS is fixed and the $V_{in}$ is varied from \SI{0.8}{\volt} to \SI{1.2}{\volt}. From \SI{0.8}{\volt} to \SI{0.9}{\volt}, the LRS shows high variability and less stability. This is due to the SET threshold not being met. On the other hand, at higher SET voltages, the variation in resistance is lower with LRS. Most importantly the SET operation range is fixed at $\sim$ \SI{4.2}{\kilo\ohm}. It indicates that without a proper pulse shape, 1T1R is not functional for SET operation.}
  \label{fig:SET_Variation_1T1R}
\end{figure}

\begin{figure}[]
  \centering
  \includegraphics[width=3.4in, height=2.9in] {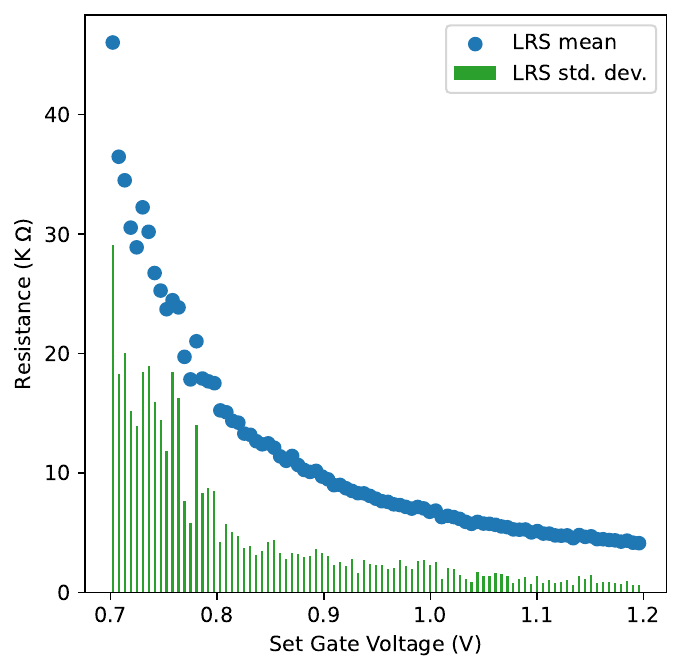}
  \caption{Measured DC test data of 3T1R configuration where MTop is fixed at a DC voltage and the gate of the $NMOS$ is varied with different SET voltage. From \SI{0.7}{\volt} to \SI{0.9}{\volt}, the LRS shows high variability and less stability. On the other hand, at higher SET voltage, the std. dev. of the resistance is lower with LRS. By controlling compliance current with the gate of $M_{N1}$ a resistance range from 5 k$\Omega$ and 20 k$\Omega$ can be targeted during programming in a DC environment. Therefore no pulse shaping is required.}
  \label{fig:SET_Variation_3T1R}
\end{figure}

\begin{figure}[]
  \centering
  \includegraphics[width=3.4in]{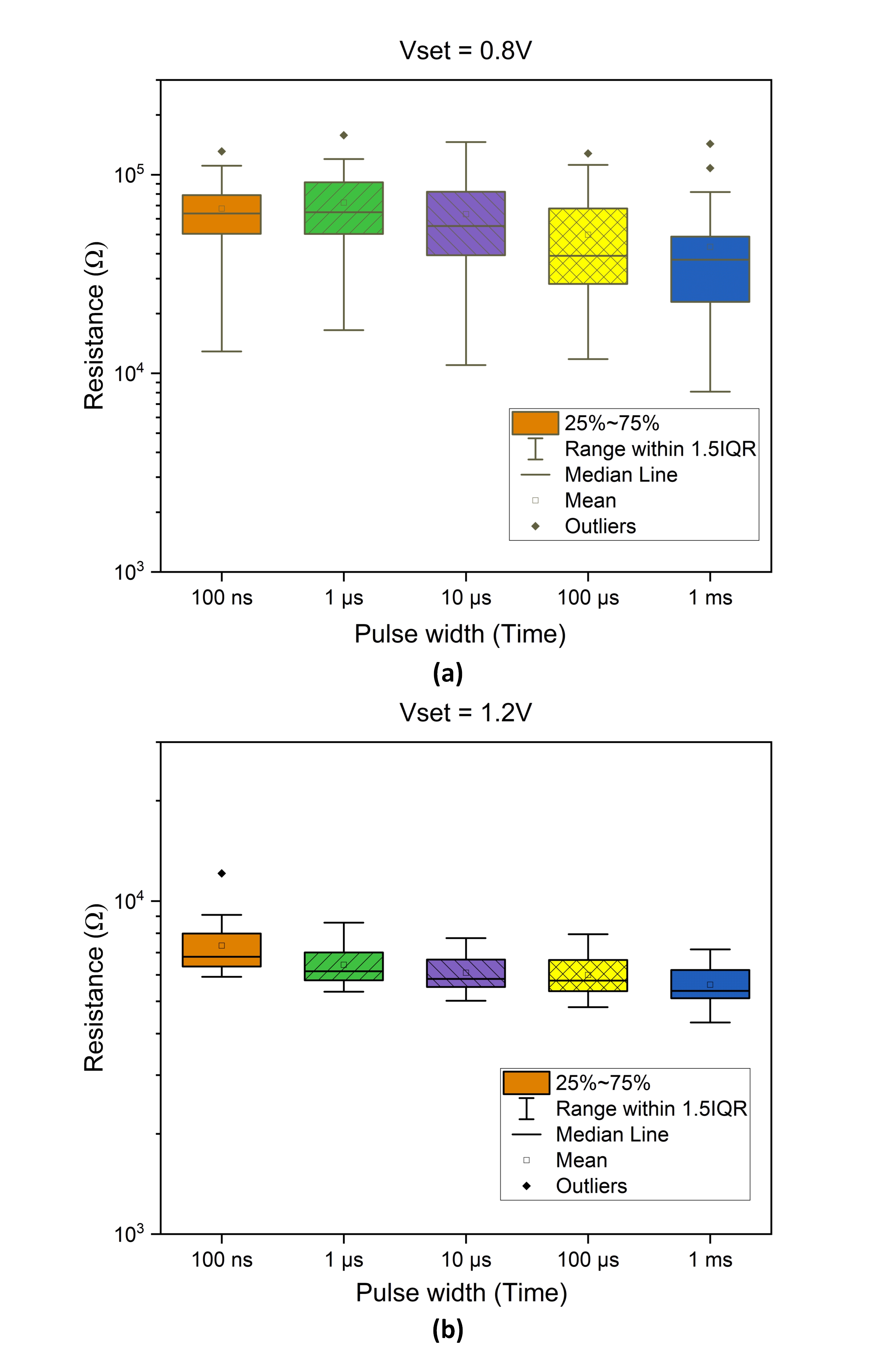}
  \caption{Measured data of 20 1T1R configurations programmed with different pulse widths. (a) shows the pulse width variation for 1T1R device configuration at \SI{0.8}{\volt}. Here the $V_{g}$ is fixed at \SI{1.2}{\volt} and the $V_{in}$ varied from lower to higher voltage pulse width. The pulse width is varied from \SI{100}{\nano\s} to \SI{1}{\milli\s}. (b) illustrates the pulse width effect on 1T1R configuration at \SI{1.2}{\volt}. Here, the pulse width variation shows a lower effect compared to the effect at \SI{0.8}{\volt}.}
  \label{fig:pulse_1T1R}
\end{figure}

1T1R and 3T1R memristive cells were fabricated as test structures using a \SI{65}{\nano\meter} CMOS technology. Hafnium oxide is used for the memristive layer. After fabrication, the raw wafer was tested in our probe station. We have set up different test cases for the devices such as forming, RESET, SET, READ, and endurance. In total, 38 devices were selected for the measurement test. Fig. \ref{fig:chip} shows our fabricated chip. There are different test structures constructed with 1T1R and 3T1R combinations. All the signal names are labeled accordingly for 1T1R and 3T1R devices.

 \subsection{Resistance Variation at LRS}
Fig. \ref{fig:SET_Variation_1T1R} exhibits the impact of DC operation. For that test, ten 1T1R cells were taken under consideration at different SET voltages. A ``KEITHLEY 2636A SYSTEM SourceMeter" is utilized to provide DC voltage for the SET operation at $V_{in}$ and $V_{g}$. Here the $V_{g}$ is fixed at \SI{1.2}{\volt} and $V_{in}$ is varied from \SI{0.8}{\volt} to \SI{1.2}{\volt}. According to Fig. \ref{fig:SET_Variation_1T1R} the std. dev. of the resistance is very high from \SI{0.8}{\volt} to \SI{0.9}{\volt}. In addition, the std. dev. is comparatively lower after \SI{0.9}{\volt}. However, the SET resistance window is limited to a narrow region around $\sim$ \SI{4.2}{\kilo\ohm}, which is not a functional SET window for our targeted programming region (\SI{5}{\kilo\ohm} to \SI{20}{\kilo\ohm}). Due to the absence of proper pulse width, the 1T1R device is not functional for SET operation.

Fig. \ref{fig:SET_Variation_3T1R} shows the LRS variation and distribution at a low $V_{in}$ from 38 experimental samples. According to Fig. \ref{fig:SET_Variation_3T1R}, the resulting resistance when the SET voltage is varied from \SI{0.7}{\volt} to \SI{1.2}{\volt}. The SET variation at LRS is drastically increased from \SI{0.7}{\volt} to \SI{0.9}{\volt}. Fig. \ref{fig:SET_Variation_3T1R} also shows the std. dev. value at different $V_{in}$ voltages during SET operation. The std. dev. value is significantly lowered at around \SI{0.8}{\volt} $V_{g}$. In addition, at LRS the resistance shows less variability and more reliable programming opportunity. It can be said from Fig. \ref{fig:SET_Variation_3T1R}, the LRS has a negligible variation for the SET operation in the range of \SI{5}{\kilo\ohm} to \SI{20}{\kilo\ohm}.

\subsection{Pulse Shape Effect Based on Measured Data}
 A Keithley B1500A semiconductor parameter analyzer was utilized to provide the specific pulse widths to the devices. Twenty devices were measured for different SET voltages at various pulse widths. Fig. \ref{fig:pulse_1T1R} (a) shows the pulse width effect at \SI{0.8}{\volt}. Lower pulse width results in higher resistance after a SET operation. Conversely, a higher pulse width results in a lower final resistance after the SET operation. In addition, the SET resistance is not in our targeted LRS region for programming. Due to that, \SI{0.8}{\volt} as a SET voltage is not reliable with various pulse widths. Fig. \ref{fig:pulse_1T1R} (b) exhibits the pulse shaping effect on the 1T1R devices at \SI{1.2}{\volt} for $V_{in}$. Here the pulse width has less effect on the SET resistance. 
 According to Fig. \ref{fig:pulse_cadence_1T1R}, there are minimal effects of pulse width on the 3T1R design. Thus, for the work shown in Fig. \ref{fig:pulse_1T1R}, the pulse width analysis is not considered for the 3T1R design. However, when compared to Fig. \ref{fig:pulse_1T1R} (a) there are some similarities that can be observed between the relationship of resistance and pulse width with different $V_{set}$ voltages. For instance, as the pulse width increases, the resistance decreases. This effect is less apparent for \SI{0.8}{\volt} since only a partial filament is formed due to the low $SET$ voltage. It can also be seen that with a lower $V_{set}$ the resulting resistance is higher overall.

\subsection{Aging and Failure Effects on the Device}
Device failure was investigated in a previous paper on 1T1R devices with respect to voltage and endurance \cite{maxage}. The paper showed that devices fail due to either current, voltage, or aging effects (endurance/cycling). If these elements are limited to certain ranges of resistance and a smaller difference between LRS and HRS is maintained, the ReRAM device can survive for a longer time. Thus, the main failure inducer is long-term switching/ endurance aging. It has been shown previously that devices can switch for over 100 million cycles. In addition, 40\% of devices can manage 1 billion cycles at room temperature \cite{maxage}. Another endurance test showed over 10 years of retention is possible at 125$^{\circ}$C temperature \cite{10years}. Device failure mitigation can be addressed throughout testing through different schemes such as the co-optimization framework \cite{age}. However, in this work device aging will be handled by adjusting the weight mapping between software to hardware based on the device's current performance.

 \section{Software Model \& Applications}
\subsection{Software Model}
A single-layer machine learning classification model can be explained as $\sigma(WX+b)$ where $\sigma$ is the activation function. The linear product ($(WX+b)$) obtains non-linearity by the activation function which is required by the machine learning model. The dot product of weight matrix W and input vector X can be calculated using the memristive crossbar dot product engine. The size of the matrix, W is the number of input features, and a number of output classes. The bias term can be mapped as the fixed column currents independent of the input as it appears from the equation \cite{bias}. The output class is the index of the column having the largest current, $class=argmax_{col}{\sigma(column\_currents)}$. This argmax function is non-differentiable and Therefore, the softmax function is used here as a pseudo replacement. The loss function used is the mean squared error of the actual class and model predicted class. In this paper, a software model is trained to determine the parameters of an abstract model where the weight values are floating point values. Later those values were mapped to the memristive conductances which were found from a Cadence Spectre simulation. A linear mapping strategy is followed to convert all the weights to memristive conductance values and all the bias terms to the bias current. For mapping, first of all, the weights were shifted to be positive values \cite{bias}. Then, the weights of range ($W_L,W_H$) are mapped to the conductance range ($G_L,G_H$). All transformations were offset by adding additional terms to the original bias current. If all the bias currents across columns are equal, then the bias term can be ignored and the linear product is the output current across the columns.

Four datasets, namely Iris, Wine, and Breast Cancer Wisconsin (Diagnostic) and banknote from the UCI repository \cite{uci} are employed for the purpose of classification in this study. To create input spikes, the data is one-hot encoded with 4 bins. For example, the iris dataset has 4 features and 4 bins for each class. Thus, the total number of features is 16. The dataset is split into 70\% training data and 30\% testing data.  One of the advantages of one-hot encoding is that it returns a specific number of spikes (the number of features) for every data instance at all columns and therefore, a specific number of the memristor is READ for any data instance. This unique feature gives advantages to simplifying the model and the mapping to the hardware.

\begin{table*}[t]

\caption{PERFORMANCE EVALUATION BETWEEN 1T1R AND 3T1R}
\centering
{%
\begin{tabular}{|l|l|l|l|l|l|l|l|l|}
\hline
{\color[HTML]{333333} Dataset} &
  {\color[HTML]{333333} \begin{tabular}[c]{@{}l@{}}Training Acc.\\ (1T1R), (\% )\end{tabular}} &
  {\color[HTML]{333333} \begin{tabular}[c]{@{}l@{}}Testing Acc.\\ (1T1R), (\%)\end{tabular}} &
  {\color[HTML]{333333} \begin{tabular}[c]{@{}l@{}}Testing / READ\\ Energy (1T1R),\\ (\SI{}{\pico\joule})\end{tabular}} &
  {\color[HTML]{333333} \begin{tabular}[c]{@{}l@{}}Training Acc.\\ (3T1R), (\%)\end{tabular}} &
  {\color[HTML]{333333} \begin{tabular}[c]{@{}l@{}}Testing Acc.\\ (3T1R), (\%)\end{tabular}} &
  \begin{tabular}[c]{@{}l@{}}Testing / READ\\ Energy (3T1R),\\ (\SI{}{\pico\joule})\end{tabular} &
  \begin{tabular}[c]{@{}l@{}}Testing Acc.\\ Degradation\\ (\%)\end{tabular} &
  \begin{tabular}[c]{@{}l@{}}Energy\\ improvement   \\ (x=times)\end{tabular} \\ \hline
{\color[HTML]{333333} Iris} &
  {\color[HTML]{333333} 91.43} &
  {\color[HTML]{333333} 88.88} &
  {\color[HTML]{333333} 449.61} &
  {\color[HTML]{333333} 85.71} &
  {\color[HTML]{333333} 88.88} &
  63.09 &
  0 &
  $\sim$7.13x \\ \hline
{\color[HTML]{333333} Wine} &
  {\color[HTML]{333333} 96.77} &
  {\color[HTML]{333333} 85.18} &
  {\color[HTML]{333333} 1446.56} &
  {\color[HTML]{333333} 96.77} &
  {\color[HTML]{333333} 81.48} &
  205.47 &
  $\sim$3.7 &
  $\sim$7.04x \\ \hline
{\color[HTML]{333333} \begin{tabular}[c]{@{}l@{}}Breast \\ Cancer\end{tabular}} &
  {\color[HTML]{333333} 98.24} &
  {\color[HTML]{333333} 93.56} &
  {\color[HTML]{333333} 2324} &
  {\color[HTML]{333333} 97.99} &
  {\color[HTML]{333333} 93.56} &
  318.46 &
  0 &
  $\sim$7.3x \\ \hline
{\color[HTML]{333333} Banknote} &
  {\color[HTML]{333333} 93.43} &
  {\color[HTML]{333333} 91.99} &
  {\color[HTML]{333333} 296.78} &
  {\color[HTML]{333333} 92.81} &
  {\color[HTML]{333333} 91.26} &
  42.39 &
  $\sim$0.73 &
  $\sim$7x \\ \hline
\end{tabular}%
}
\label{tab:comp}
\end{table*}

\subsection{Network Architecture}
The network architecture consists of an input layer and an output layer, without any hidden layers. Initially, the trained network includes bias values. However, during the process of mapping to hardware, the bias values were adjusted to be equal. This eliminates the need for them to be represented in hardware, as detailed in \cite{bias}. TABLE \ref{tab:tab2} presents the models for the four different classifications. As an example, the iris dataset, which comprises 4 features, has each feature encoded with 4 bins, resulting in a total of 16 units in the input layer. Including the bias term, the input layer consists of 17 units. The number of units in the input layer is equivalent to the number of classes in the classification. The trainable parameters represent the parameters of the model that underwent training.

\begin{table}[]
\caption{Architecture of the Network}
\centering
{%
\begin{tabular}{|l|l|l|l|l|}
\hline
Dataset &
  \begin{tabular}[c]{@{}l@{}}Number of \\ Features\end{tabular} &
  \begin{tabular}[c]{@{}l@{}}Units in \\ Input Layer\end{tabular} &
  \begin{tabular}[c]{@{}l@{}}Units in Output\\  Layer, Number\\  of class\end{tabular} &
  \begin{tabular}[c]{@{}l@{}}Trainable\\  parameters\end{tabular} \\ \hline
Iris                                                                & 4  & 17  & 3 & 51  \\ \hline
Wine                                                                & 13 & 53  & 3 & 159 \\ \hline
\begin{tabular}[c]{@{}l@{}}Breast Cancer \\ Diagnostic\end{tabular} & 30 & 121 & 2 & 242 \\ \hline
Banknote                                                            & 4  & 17  & 2 & 34  \\ \hline
\end{tabular}%
}
\label{tab:tab2}
\end{table}

\begin{figure}[]
  \centering
  \includegraphics[width=3.4in]{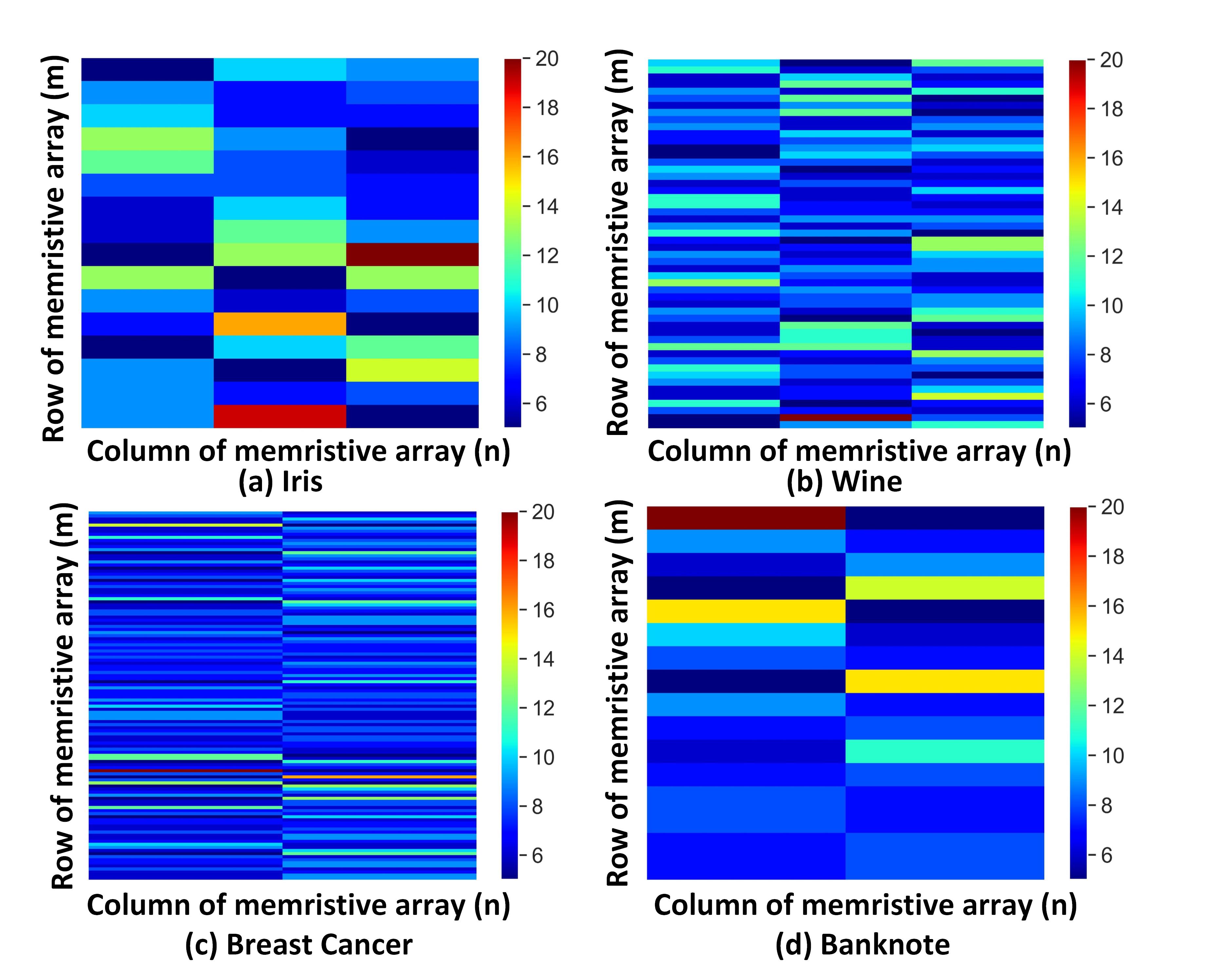}
  \caption{Model of four classification tasks. A model shows the distribution of the memristor's weight for different classification tasks such as  (a) Iris, (b) Wine, (c) Breast Cancer, and (d) Banknote. The model adapts lower weight value more to ignore the sensitivity effects of HRS.}
  \label{fig:model}
\end{figure}

\begin{figure}[]
  \centering
  \includegraphics[width=3.4in]{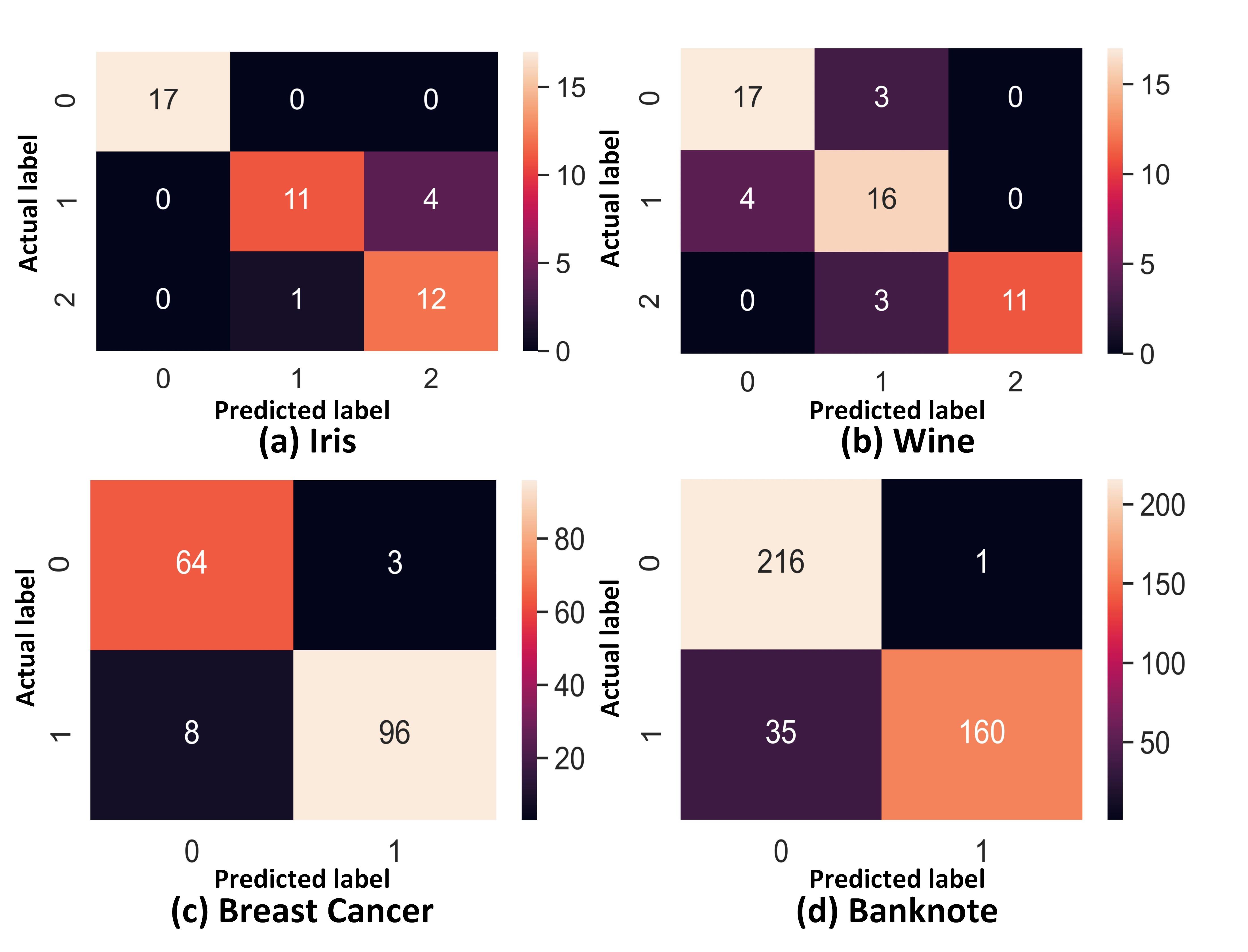}
  \caption{Confusion matrix of four datasets are shown here. Four confusion matrices (a) Iris, (b) Wine, (c) Breast Cancer, (d) Banknote illustrate the prediction efficiency of the model. If the diagonal boxes contain a maximum number of rows, the model prediction is better.  All confusion matrices show excellent prediction capability.}
  \label{fig:CM}
\end{figure}

\subsection{Results from the Software Model}
For mapping an abstract software model to the existing hardware design, a lookup table is required that provides the corresponding current value for all memristive conductances. In this work, two look-up tables are created for 1T1R and 3T1R, respectively, using the results from the Cadence Spectre simulation. In addition, the column current of the DPE is calculated using this look-up table. At the output, $n$ current levels ($I_1,I_2,....,I_n$) need to be distinguishable in order to determine the column which has the maximum current. Otherwise, all current appears to be the same due to the limitations of current sensing. In this work, it is assumed that the ADC has \SI{50}{\nano\ampere} current resolution. One of the limitations of analog DPE-based classification systems is the possibility of multiple columns sharing the maximum current. This is a limitation imposed by the ADC. In those cases, the classifier will take the class randomly from one of those columns. 

The software model also behaves similarly. Table \ref{tab:comp} shows the comparison of the DPE-based classification using 1T1R and 3T1R for four datasets. The validation metrics are training accuracy, testing accuracy, and mean energy required for the classification of a single sample. The mean energy is proportional to the size of the input vector and the memristive DPE size. Breast cancer has the largest DPE and therefore requires the largest energy among all four classification tasks. The testing accuracy degradation column shows the 3T1R-based DPE performs slightly worse in the wine dataset compared to 1T1R based DPE. But, the 3T1R has a low current range for \SI{5}{\kilo\ohm} to \SI{20}{\kilo\ohm} and therefore, suffers from current sensing limitations. A \SI{1}{\micro\s} pulse is provided at the gate of MN1 with \SI{0.6}{\volt} amplitude. Hence, a READ current is calculated from the $2^{nd}$ stage of the proposed 3T1R synapse.  The percentage of samples in each dataset in which multiple columns have at or over the maximum current sensable is described next. For the Iris, Wine, and Breast Cancer Wisconsin (Diagnostic), and Banknote dataset, the percentage same current level for 3T1R was 24.44\%, 11.11\%, 0.58\%, and 1.70\% respectively. However, for the 1T1R these results are just 4.4\%,0\%, 0\% and 0\% for the same test. This limitation of 3T1R can be overcome by using an ADC with better current sensing capability. On the other hand, the energy is improved by around $7 \times$ when using 3T1R compared to 1T1R.

In addition, in TABLE \ref{tab:comp} the testing dataset accuracy for the Wine dataset is 85.18\% when the memristor crossbar is made by the 1T1R devices. The accuracy reduces by approximately 3.7\% when 1T1R in the crossbar is replaced by 3T1R.   

The 3T1R-based trained DPE model is shown in Fig. \ref{fig:model}. The DPE model is just the resistance of the memristor which is shown as the color bar. In the color bar, the range 5-20 represents \SI{5}{\kilo\ohm}-\SI{20}{\kilo\ohm}. Lastly, Fig. \ref{fig:CM} shows the confusion matrix of the testing dataset for all four classifications. It is clear from the confusion matrix that all the models classify all the classes pretty well. 
Moreover, in this work, our targeted programming region is from \SI{5}{\kilo\ohm} to \SI{20}{\kilo\ohm}.  In the 3T1R, depending on the programming / SET voltage, sixteen different resistance levels are targeted. In prior work, it has been shown up to 16 levels of resistance can be achieved with our memristive device \cite{maxage}. In the DPE configuration, if the array is small then the cumulative current in a column shows almost negligible degradation in precision. However, if the column is containing more than 16 synaptic units then the cumulative current shows saturation in the column current.

\section{Peripheral Circuitry}
A CMOS digital-to-analog converter (DAC) is utilized for programming. The energy consumption of our DAC to program the synapse with 1V is \SI{0.311}{\femto\joule}. Here, \SI{1}{\volt} is the output of the DAC circuit from \SI{1.2}{\volt} as the supply voltage.  Moreover, a Winner-Take-All (WTA) circuit is utilized to rank the column current from the cross-bar array. With a particular sample from the Wine dataset, the column current from each column is about \SI{11.2}{\micro\ampere}, \SI{11.5}{\micro\ampere}, and \SI{11.3}{\micro\ampere} for $Iout1$, $Iout2$, and $Iout3$ respectively. Our designed WTA circuit consumes \SI{0.607}{\pico\joule} energy with \SI{1}{\micro\s} read pulse to detect the maximum current from three columns. Three columns need only nine $NMOS$ transistors to classify the maximum current. Hence, a sense amplifier is used to convert the analog value into a digital value with one bit per column.

\section{Comparison with prior work}

An analog implementation of a DPE can consume less power during each READ cycle compared to traditional memory like SRAM and DRAM \cite{dm2, dm3, ref6}. According to TABLE \ref{tab:Comparison_prior},  in \cite{b1}, authors utilized SRAM with 4-bit precision at \SI{45}{\nano\meter} CMOS process to implement a DPE and it consumed \SI{120}{\micro\ampere} for reading at \SI{0.4}{\volt}. Whereas a memristive analog memory implementation is used to implement a DPE and it consumed only \SI{27.58}{\micro\ampere} at \SI{2}{\volt} \cite{ref5}. In our paper, two different approaches are observed for a DPE implementation. The 1T1R approach draws \SI{83.52}{\micro\ampere} current at \SI{0.6}{\volt}. On the other hand, the 3T1R approach shows, \SI{4.459}{\micro\ampere} of READ current at \SI{0.6}{\volt}. Our proposed 3T1R device shows about $\sim$26.91x and $\sim$6.19x READ current improvement compared to \cite{b1} and \cite{ref5} respectively. In addition, the 3T1R approach can reduce $\sim$18.73x READ current compared to a 1T1R design. Another research group presents a mixed-signal synapse circuit with 5-bit data precision\cite{ref2}. The READ power consumption of this design is \SI{27.43}{\micro\watt}, which is $\sim$5.15x higher than our 3T1R approach. Another memristive-based design draws \SI{14}{\micro\watt} READ power, which is $\sim$2.62x higher than our proposed 3T1R design\cite{ref6}. Moreover, the 3T1R design consumes $\sim$9.37x lower READ power compared to the 1T1R design. 
 A memristor based design with \SI{28}{\nano\meter} CMOS process shows \SI{17.8}{\pico\joule} energy consumption\cite{ref4} which has $\sim$ 3.33x energy overhead compared to our 3T1R design. Moreover, the 3T1R design can save $\sim$ 9.37x energy compared with the 1T1R design. 
However, according to Fig. \ref{fig:IP} (c) and (d), 3T1R shows about 2x and 6x \textit{SET}  current and power overhead compared to the 1T1R structure. This higher \textit{SET} current can be reduced by lowering the memristive device's overall  \textit{SET} threshold and using a thin oxide transistor instead of a thick oxide transistor, which is targeted for future work.
 The overall area of a 3T1R synapse is almost 2.76x larger than a 1T1R synapse. However, the 3T1R design shows significant area improvement compared to prior work\cite{ref2}. The area overhead of a 3T1R-based DPE can be optimized further by sharing the  $MP3$ for a whole column. In Fig. \ref{fig:layout}, each synapse utilizes it's own $MP3$ transistor.

\begin{table*}[t]
\centering
    \caption{READ PERFORMANCE COMPARISON WITH PRIOR WORK}

{%
\begin{tabular}{|l|l|l|l|l|l|l|l|}
\hline
{\color[HTML]{333333} Ref.} &
  {\color[HTML]{333333} \cite{b1}} &
  {\color[HTML]{333333} \cite{ref2}} &
  {\color[HTML]{333333} \cite{ref5}} &
  {\color[HTML]{333333} \cite{ref4}} &
  {\color[HTML]{333333} \cite{ref6}} &
  {\color[HTML]{333333} \begin{tabular}[c]{@{}l@{}}This work\\ (1T1R)\end{tabular}} &
  \begin{tabular}[c]{@{}l@{}}This work  \\ (3T1R)\end{tabular} \\ \hline
{\color[HTML]{333333} CMOS Process} &
  {\color[HTML]{333333} 45nm} &
  {\color[HTML]{333333} 180nm} &
  {\color[HTML]{333333} -} &
  {\color[HTML]{333333} 28nm} &
  {\color[HTML]{333333} -} &
  {\color[HTML]{333333} 65nm} &
  65nm \\ \hline
{\color[HTML]{333333} Resolution} &
  {\color[HTML]{333333} 4-bit} &
  {\color[HTML]{333333} 5-bit} &
  {\color[HTML]{333333} -} &
  {\color[HTML]{333333} 2-bit} &
  {\color[HTML]{333333} -} &
  {\color[HTML]{333333} 4-bit} &
  4-bit \\ \hline
{\color[HTML]{333333} Device} &
  {\color[HTML]{333333} SRAM} &
  {\color[HTML]{333333} -} &
  {\color[HTML]{333333} Memristor} &
  {\color[HTML]{333333} Memristor} &
  {\color[HTML]{333333} Memristor} &
  {\color[HTML]{333333} Memristor} &
  Memristor \\ \hline
{\color[HTML]{333333} Material} &
  {\color[HTML]{333333} CMOS} &
  {\color[HTML]{333333} CMOS} &
  {\color[HTML]{333333} -} &
  {\color[HTML]{333333} \ce{HZO}} &
  {\color[HTML]{333333} \ce{HfO2}} &
  {\color[HTML]{333333} \ce{HfO2}} &
  HfO2 \\ \hline
{\color[HTML]{333333} READ voltage} &
  {\color[HTML]{333333} 0.4V} &
  {\color[HTML]{333333} -} &
  {\color[HTML]{333333} 2V} &
  {\color[HTML]{333333} -} &
  {\color[HTML]{333333} 1.25V} &
  {\color[HTML]{333333} 0.6V} &
  0.6V \\ \hline
{\color[HTML]{333333} READ current} &
  {\color[HTML]{333333} \SI{120}{\micro\ampere}} &
  {\color[HTML]{333333} -} &
  {\color[HTML]{333333} \SI{27.58}{\micro\ampere}} &
  {\color[HTML]{333333} -} &
  {\color[HTML]{333333} -} &
  {\color[HTML]{333333} \begin{tabular}[c]{@{}l@{}}\SI{83.52}{\micro\ampere} \\ at \SI{5}{\kilo\ohm}\end{tabular}} &
  \begin{tabular}[c]{@{}l@{}}\SI{4.459}{\micro\ampere} \\ (Iin+I2) \\ Fig. \ref{fig:3T1R} (b)\end{tabular} \\ \hline
READ power &
  - &
  \SI{27.43}{\micro\watt} &
  - &
  - &
  \SI{14}{\micro\watt} &
  \SI{50.11}{\micro\watt} &
  \SI{5.35}{\micro\watt} \\ \hline
READ energy &
  - &
  - &
  - &
  \SI{17.8}{\pico\joule} &
  - &
  \SI{50.11}{\pico\joule} &
  \SI{5.35}{\pico\joule} \\ \hline
\begin{tabular}[c]{@{}l@{}}READ current\\ improvement\end{tabular} &
  \textbf{$\sim$26.91x} &
   &
  \textbf{$\sim$6.19x} &
  - &
  - &
  \textbf{$\sim$18.73x} &
  \textbf{1x} \\ \hline
\begin{tabular}[c]{@{}l@{}}READ power\\ improvement\end{tabular} &
  - &
  \textbf{$\sim$5.15x} &
  - &
  - &
  \textbf{$\sim$2.62x} &
  \textbf{$\sim$9.37x} &
  \textbf{1x} \\ \hline
  \begin{tabular}[c]{@{}l@{}}READ energy\\ improvement\end{tabular} &
  - &
  - &
  - &
  - &
  \textbf{$\sim$3.33x} &
  \textbf{$\sim$9.37x} &
  \textbf{1x} \\ \hline
  \begin{tabular}[c]{@{}l@{}}Area Evaluation\end{tabular} &
  - &
 \textbf{$\sim$56x improvement} &
  - &
  - &
  \textbf{-} &
  \textbf{$\sim$2.76x overhead} &
  \textbf{1x} \\ \hline
   \begin{tabular}[c]{@{}l@{}}SET Current \\ Evaluation\end{tabular} &
  - &
 \textbf{-} &
  - &
  - &
  \textbf{-} &
  \textbf{$\sim$2x overhead} &
  \textbf{1x} \\ \hline
   \begin{tabular}[c]{@{}l@{}}SET Power\\ Evaluation\end{tabular} &
  - &
 \textbf{-} &
  - &
  - &
  \textbf{-} &
  \textbf{$\sim$6x overhead} &
  \textbf{1x} \\ \hline
\end{tabular}%

}
\label{tab:Comparison_prior}
\end{table*}

\section{Conclusion and future work}

Array-based memristive memory for SNNs are power-hungry during READ cycles. They also show high sensitivity due to process variation in MOSFET fabrication and the inherent stochastic behavior of memristors. In this work, we leveraged a 3T1R design to achieve a reduction in  READ current and process variation sensitivity. According to simulation results, $\sim$26.91x and $\sim$18.76x of the READ-out current reduction can be achieved as compared to the prior work and 1T1R approach respectively. In addition, both READ power and energy are saved by about 9.37x with the 3T1R device when compared with the 1T1R device. At the same time, the write variation is also reduced by targeting a programmable LRS, ranging from 5 k$\Omega$ to 20 k$\Omega$. This region was chosen since it contains the least variability in resistance during programming. Finally, hardware-realistic Cadence Spectre results were used to verify a software model for four different dataset classification tasks. Due to the small change in READ current across the resistance range of our 3T1R DPE, its precision is impacted. As a result, classification tasks show a slight decrease in testing accuracy compared to the 1T1R-based DPE. Overall, the proposed design is power efficient and reliable. For future work, the SET currents of the 3T1R will be optimized alongside further testing of our fabricated chip. At the same time, the output precision will be enhanced for the 3T1R cells. Optimized electrical processes such as write-read-verify and ultra-fast pulsing, pulses less than \SI{10}{\nano\s}, can be used to achieve more distinct resistance levels. Further analysis of scaling issues due to peripheral circuitry and interconnect distances will be taken under consideration. In addition, a super-resolution crossbar can be a potential candidate to explore and compare with our design \cite{S1,S2}.       


\bibliographystyle{IEEEtran}
 \bibliography{bibliography}

\vspace{11pt}
\bf{}\vspace{-33pt}
\begin{IEEEbiography}[{\includegraphics[width=1in,height=1.25in,clip,keepaspectratio]{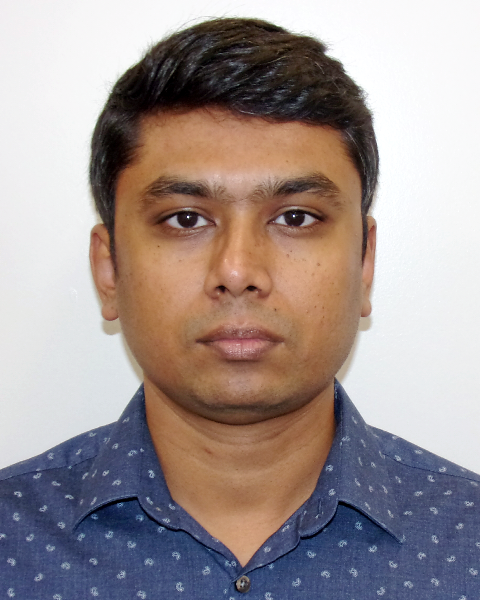}}]{Hritom Das} (Member, IEEE) received the B.Sc. degree in electrical and electronic engineering from American International University-Bangladesh, Dhaka, Bangladesh, the M.Sc. degree in electronic engineering from Kyungpook National University, Daegu, South Korea, and the Ph.D. degree in electrical and computer engineering from North Dakota State University, Fargo, North Dakota, in 2012, 2015, and 2020 respectively. He was a visiting Assistant Professor with the department of electrical and computer engineering at the University of South Alabama, Mobile, AL, USA. Currently, he is a Post-Doctoral Research Associate with the department of electrical engineering and computer science at The University of Tennessee, Knoxville, TN, USA. His research interest includes low-power VLSI circuit design, optimization, and testing. He is also exploring machine learning implementation on traditional electronics.  In addition, he is working on neuromorphic system design and optimization.
\end{IEEEbiography}

 \vskip 0pt plus -1fil
 
\begin{IEEEbiography}[{\includegraphics[width=1in,height=1.25in,clip,keepaspectratio]{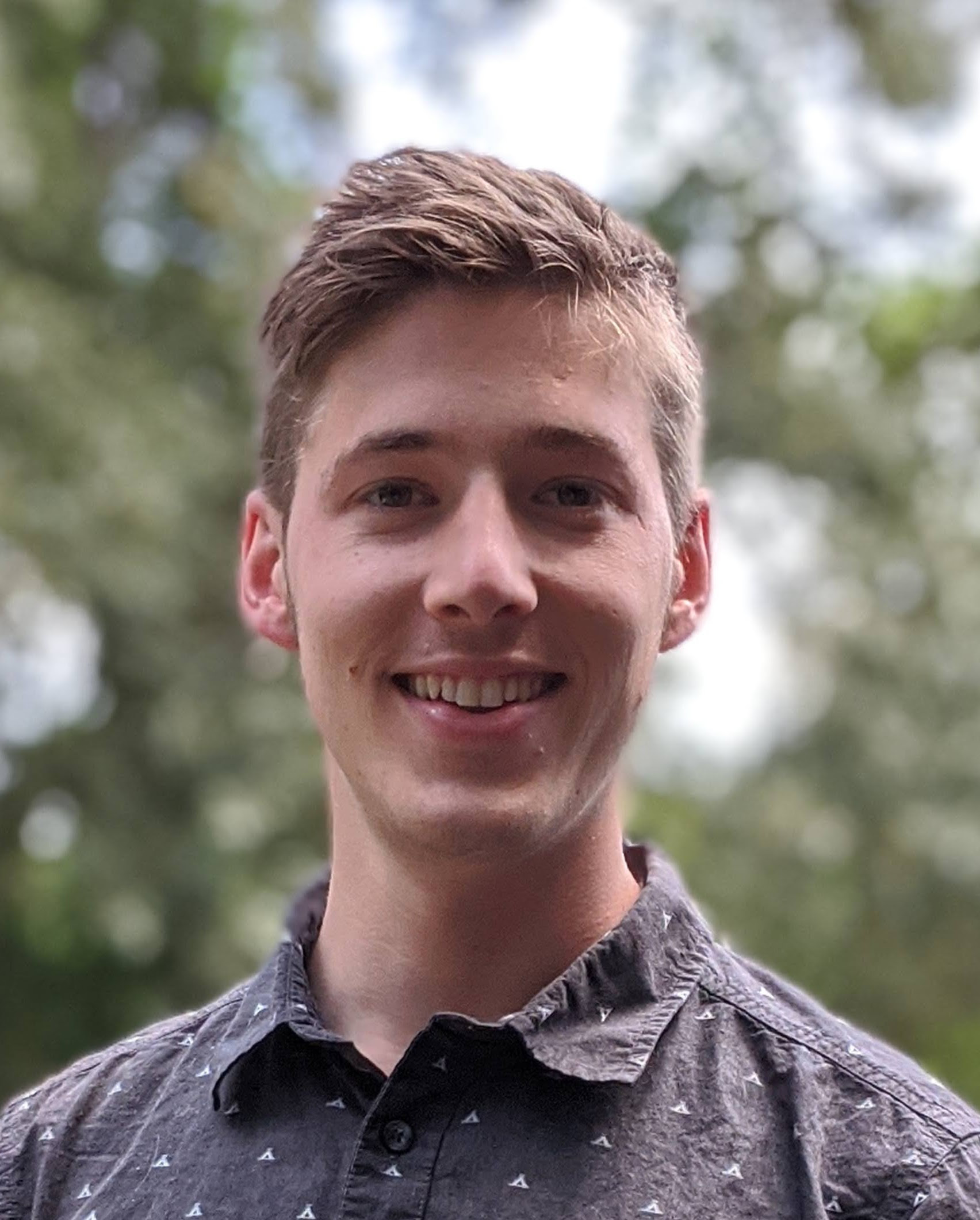}}] {Rocco Febbo} is currently enrolled at University of Tennessee, Knoxville (UTK) as a Ph.D student. He received his B.S in electrical engineering from UTK in 2020. His research interests include working with memristive devices to design and test low power neuromorphic circuits as well as designing digital systems for neuromorphic computing.
\end{IEEEbiography}

\vskip 0pt plus -1fil

\begin{IEEEbiography}[{\includegraphics[width=1in,height=1.25in,clip,keepaspectratio]{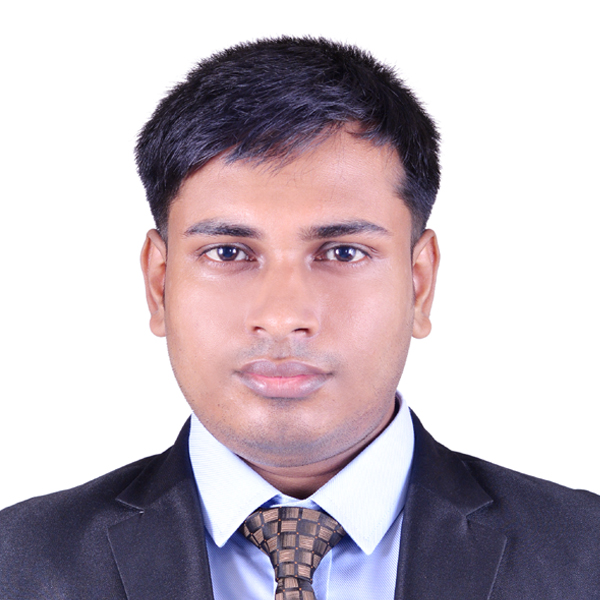}}] {SNB Tushar} (Graduate Student Member, IEEE) is currently enrolled at University of Tennessee, Knoxville (UTK) as a Ph.D student. He received his B.S in electrical and electronic engineering from Chittagong University of Engineering and Technology and MS in electrical engineering from Texas State University, San Marcos. His research interests include implementations of machine learning and neuromorphic computing using the memristive device.
\end{IEEEbiography}

 \vskip 0pt plus -1fil

\begin{IEEEbiography}[{\includegraphics[width=1in,height=1.25in,clip,keepaspectratio]{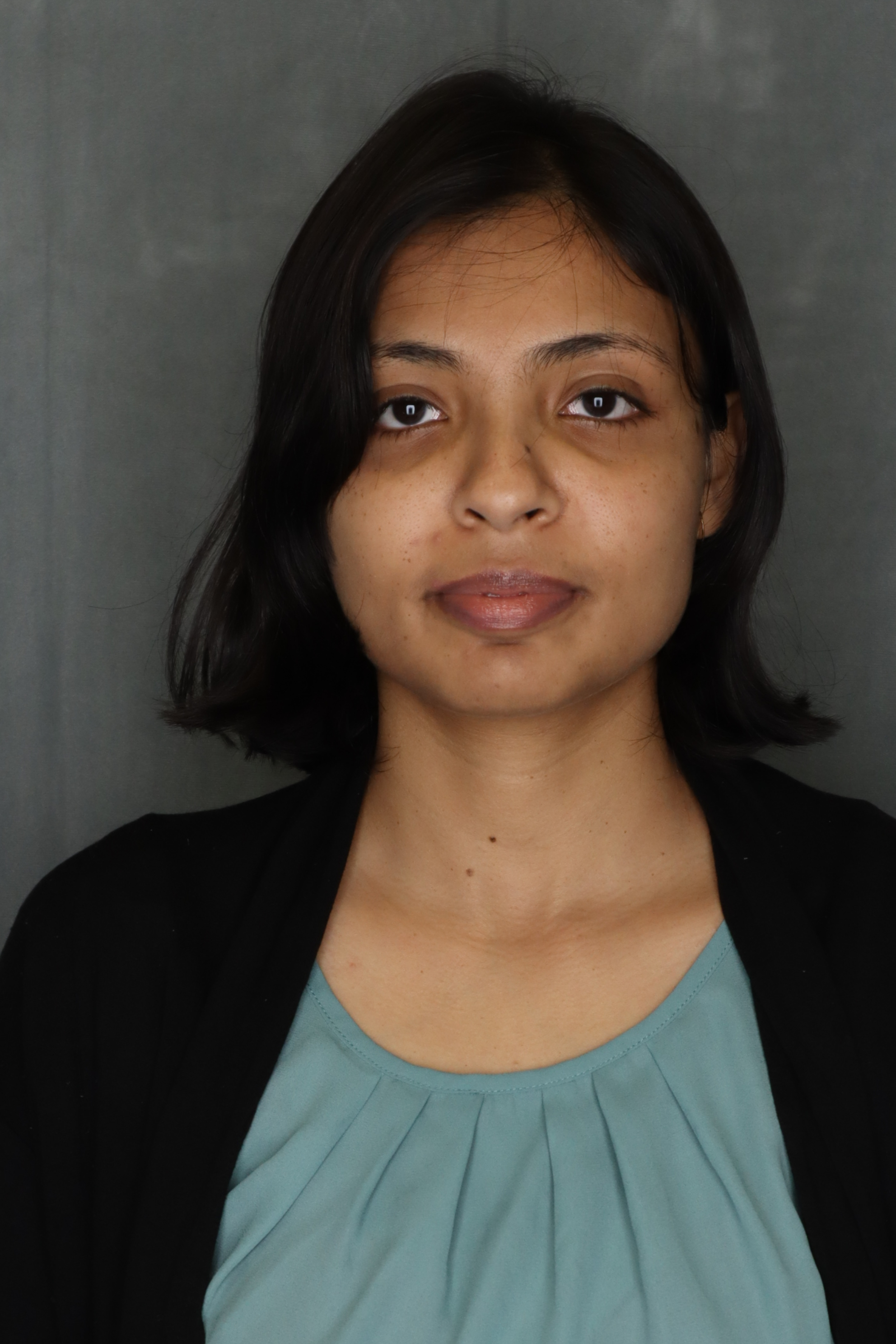}}] {Nishith N. Chakraborty} (Graduate Student Member, IEEE) is currently enrolled at the University of Tennessee, Knoxville (UTK) as a Ph.D student. She received her B.Sc. degree in electrical and electronic engineering from Bangladesh University of Engineering and Technology, and MS in electrical engineering from University of California, Riverside. Her research interest includes analog mixed-signal neuromorphic circuits design and optimization, low-power VLSI circuits and memristor-based circuit design.
\end{IEEEbiography}

\vskip 0pt plus -1fil

\begin{IEEEbiography}[{\includegraphics[width=1in,height=1.25in,clip,keepaspectratio]{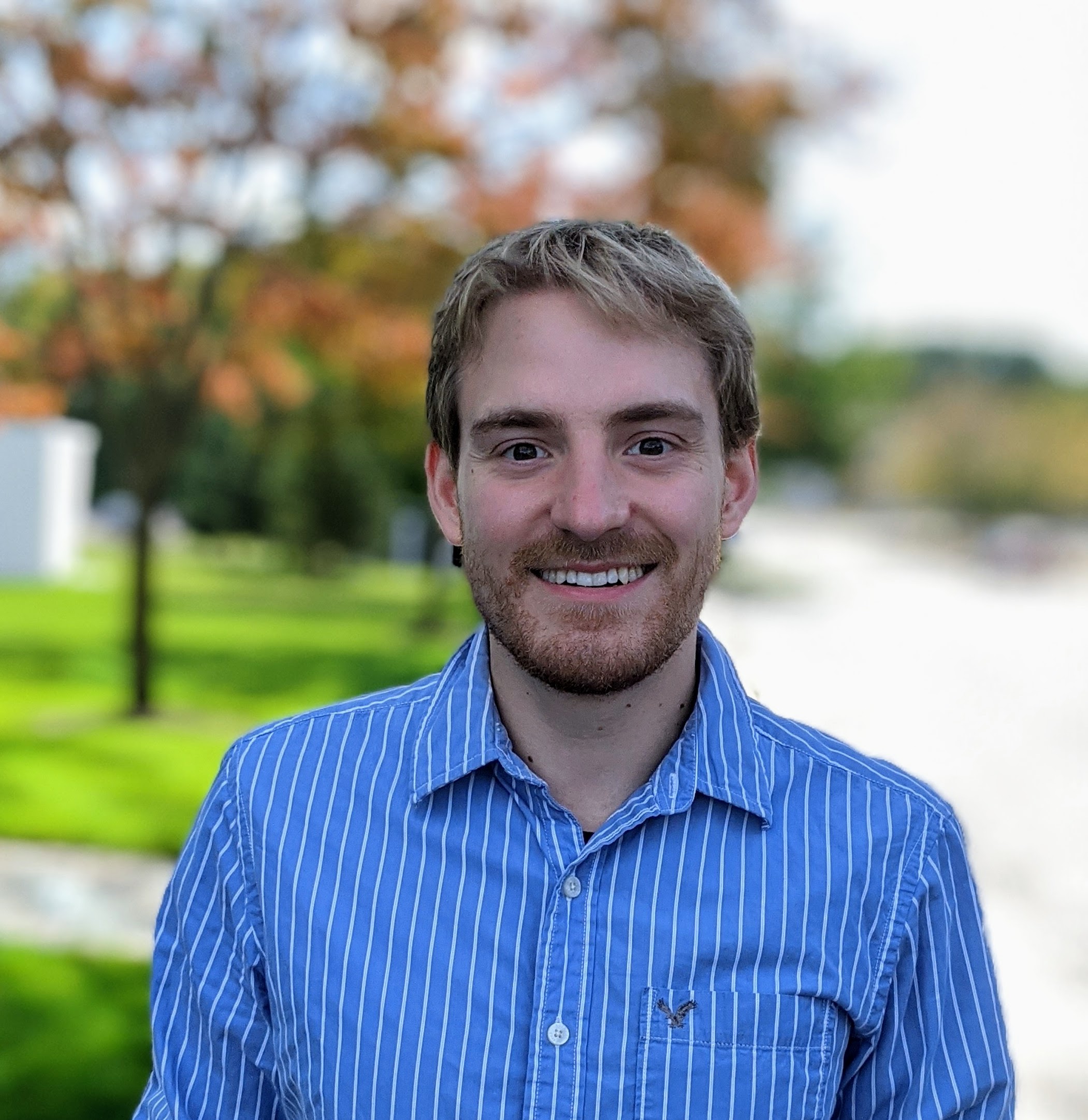}}] {Maximilian Liehr} (Member, IEEE) obtained his BA and MEng from Rensselaer Polytechnic Institute in Troy, NY and his Ph D. from the Colleges of Nanoscale Science \& Engineering at SUNY Polytechnic Institute in Albany, NY. He is currently a Post-Doctoral Research Associate at SUNY Polytechnic Institute. He has active research interests in development of Non-volatile nanoelectronics including Resistive Random Access Memory (ReRAM) devices and neuromorphic computing.
\end{IEEEbiography}

\vskip 0pt plus -1fil

\begin{IEEEbiography}[{\includegraphics[width=1in,height=1.25in,clip,keepaspectratio]{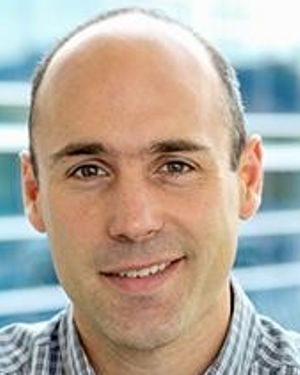}}] {Nathaniel Cady} (Member, IEEE) Prof. Cady earned his BA and Ph.D. from Cornell University in Ithaca, NY and is currently an Empire Innovation Professor of Nanobioscience and Interim Vice President of Research in the College of Nanoscale Science \& Engineering at SUNY Polytechnic Institute. Prof. Cady has active research interests in the development of novel biosensor technologies and biology-inspired nanoelectronics, including novel hardware for neuromorphic computing. He is also the executive director of the SUNY Applied Materials Research Institute (SAMRI) that funds collaborative research efforts between SUNY faculty and industry partner Applied Materials (AMAT). 
\end{IEEEbiography}

\vskip 0pt plus -1fil

\begin{IEEEbiography}[{\includegraphics[width=1in,height=1.25in,clip,keepaspectratio]{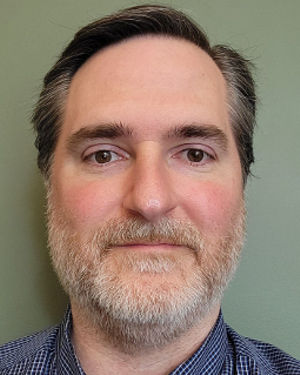}}] {Garrett S. Rose} (Senior Member, IEEE) received
the B.S. degree in computer engineering from Virginia Polytechnic Institute and State University (Virginia Tech), Blacksburg, VA, USA, in 2001, and the M.S. and Ph.D. degrees in electrical engineering from the University of Virginia, Charlottesville, VA, USA,in 2003 and 2006, respectively. His Ph.D. dissertation was on the topic of circuit design methodologies for molecular electronic circuits and computing architectures. He is currently a Professor with the Min H. Kao Department of Electrical Engineering and Computer Science, University of Tennessee, Knoxville, TN, USA, where his work is focused on research in the areas of nanoelectronic circuit design, neuromorphic computing, and hardware security. From June 2011 to July 2014, he was with the Air Force Research Laboratory, Information Directorate, Rome, NY, USA. From August 2006 to May 2011, he was an Assistant Professor with the Department of Electrical and Computer Engineering, Polytechnic Institute of New York University, Brooklyn, NY, USA. From May 2004 to August 2005, he was with MITRE Corporation, McLean, VA, USA, involved in the design and simulation of nanoscale circuits and systems. His research interests include low-power circuits, system-on-chip design, trusted hardware, and developing VLSI design methodologies for novel nanoelectronic technologies.
\end{IEEEbiography}

\end{document}